\pdfoutput=1
\documentclass[lettersize,journal]{IEEEtran}
\usepackage{amsmath,amsfonts}
\usepackage{array}
\usepackage[caption=false,font=normalsize,labelfont=sf,textfont=sf]{subfig}
\usepackage{textcomp}
\usepackage{stfloats}
\usepackage{url}
\usepackage{verbatim}
\usepackage{graphicx}
\usepackage{balance}
\usepackage[colorlinks=true, urlcolor=blue]{hyperref}        
\usepackage{booktabs}
\usepackage{multirow}
\usepackage{xcolor}
\usepackage{framed}
\usepackage{soul}
\usepackage{makecell}

\begin{document}

\title{When Misalignment Becomes Supervision: Structured Label Noise in Supervised Synthetic CT Generation}

\author{Valentin Boussot$^{\dagger}$, Cédric Hémon$^{\dagger}$, Caroline Lafond, Jean-Claude Nunes, Jean-Louis Dillenseger
\thanks{V. Boussot, C. Hémon, C. Lafond, J-C. Nunes and J-L. Dillenseger are with Univ. Rennes, CLCC Eugene Marquis, INSERM, LTSI - UMR 1099, F-35000 Rennes, France. Corresponding author: V. Boussot (e-mail: boussot.v@gmail.com).\\
$^{\dagger}$These authors contributed equally to this work.}}

\maketitle

\begin{abstract}

Supervised synthetic CT generation is commonly formulated as a voxel-wise regression problem between MRI or CBCT inputs and registered reference CT images. This formulation assumes that paired images are spatially aligned at the voxel level, although in practice multimodal pairs are constructed through registration procedures that inevitably leave residual anatomical misalignments. These residual errors are not independent intensity noise, but spatially coherent geometric discrepancies that can act as structured label noise during training.

In this work, we investigate how registration-induced supervision bias affects supervised MRI-to-CT and CBCT-to-CT synthesis. We show that quantitative performance strongly depends on the consistency between the registration strategy used to construct the training targets and the one used during evaluation. Models achieve better voxel-wise scores when both conventions match, indicating that supervised synthesis networks can partially learn the geometric convention imposed by the registration pipeline. This effect also impacts out-of-distribution robustness and predictive uncertainty, with less geometrically consistent supervision leading to larger prediction variability.

This work does not aim to solve registration errors through a new synthesis architecture, but to demonstrate that residual registration defines a supervision convention that supervised networks can learn and that standard metrics can reward. We validate this analysis on 1,784 paired patients from six clinical centers, covering MRI-to-CT and CBCT-to-CT synthesis across five anatomical regions. To mitigate the limitations of purely voxel-wise supervision, we introduce a SAM-based perceptual loss that compares synthesized and reference CT images in the feature space of a pretrained Segment Anything encoder. Compared with MAE-only and VGG-based perceptual objectives, this supervision improves downstream anatomical metrics and produces sharper, more structurally coherent synthetic CT images. We further show that perceptual and voxel-wise metrics may disagree when references are imperfectly aligned, while becoming more consistent when the evaluation geometry is reliable.

Overall, this study identifies registration-induced bias as a central confounder in supervised synthetic CT generation. Our results argue that
synthetic CT methods should not be evaluated solely through voxel-wise agreement with imperfect reference images, but also through anatomy-oriented criteria that assess whether patient-specific structures are faithfully preserved.
\end{abstract}


\providecommand{\mstd}[3]{\shortstack{#1$^{#3}$\\[0.4ex]{\scriptsize$\pm$#2}}}
\providecommand{\bmstd}[3]{\shortstack{\textbf{#1}$^{#3}$\\[0.4ex]{\scriptsize\textbf{$\pm$#2}}}}

\section{Introduction}

Synthetic CT generation aims to transform MRI or CBCT images into CT-like images that can be used by downstream tools originally designed for CT images, including dose calculation, segmentation, and image registration \cite{edmund2017review,dayarathna2024deep}. In recent years, supervised image-to-image translation has become the dominant paradigm for this task, largely driven by public benchmarks and large paired datasets. In this setting, models are trained to minimize voxel-wise reconstruction losses between the synthesized CT and a reference CT, and performance is commonly assessed using intensity-based metrics such as MAE, PSNR, and SSIM.

This indicator-based approach raises a Goodhart-type concern: when an indicator becomes the target of optimization, it may no longer be a reliable indicator of the underlying objective it was intended to measure \cite{goodhart1975problems}. In synthetic CT image generation, the goal is not simply to reproduce a reference image voxel by voxel, but to produce a CT-like image that preserves the patient-specific anatomy of the source modality. If the reference used for supervision and evaluation is imperfect, optimizing voxel-level scores may therefore prioritize consistency with the reference’s construction process rather than anatomical fidelity.

This issue is particularly relevant because the supervised formulation relies on a strong geometric assumption: the input image and the reference CT must be spatially aligned at the voxel level. In practice, MRI-CT and CBCT-CT pairs are acquired at different times, under different acquisition conditions, and often in different anatomical configurations \cite{edmund2017review,nyholm2018mr}. The reference CT used for supervision is therefore not a true voxel-wise ground truth, but a registered target that may contain residual alignment errors. These residuals do not behave as independent intensity noise. They correspond to spatially coherent anatomical displacements and can therefore be interpreted as structured geometric label noise \cite{florkow2019impact,kong2021breaking,dayarathna2024deep}.

This has important consequences for both training and evaluation. During training, voxel-wise losses encourage the model to reproduce the statistical structure of the registered targets, including residual deformations and systematic biases introduced by the registration pipeline \cite{zhou2023mitigating,li2025boosting,lee2025meta}. During evaluation, reference-based metrics may reward agreement with a particular registration convention rather than faithful preservation of the source anatomy \cite{dohmen2024similarity}. As a result, a model may achieve strong quantitative scores while partially learning or reproducing misalignment patterns embedded in the supervised data.

In this work, we investigate registration-induced supervision bias in supervised synthetic CT generation. Our primary objective is to determine whether residual registration defines a geometric convention that synthesis networks can learn and that reference-based metrics can subsequently reward. We analyze how the registration method used to construct training and evaluation pairs affects quantitative performance, out-of-distribution (OOD) behavior, and predictive uncertainty \cite{hemon2026towards}. This analysis tests whether measured performance reflects synthesis fidelity alone or also agreement with the registration pipeline.

To mitigate the limitations of purely voxel-wise supervision, we further introduce a SAM-based perceptual loss for synthetic CT generation. Instead of comparing images only at the intensity level, this loss compares synthesized and reference CT images in the feature space of a pretrained Segment Anything encoder \cite{kirillov2023segment}. It is intended to promote spatially coherent anatomical structures and reduce regression-induced blurring, but it does not eliminate the systematic geometric bias contained in imperfectly registered training targets.

Overall, this study argues that synthetic CT generation should not be evaluated solely as an intensity regression problem. When reference images are imperfectly aligned, voxel-wise metrics can become confounded by registration errors and may reward conformity to a registration convention rather than preservation of the source anatomy. SAM-based perceptual supervision is therefore investigated as a partial anatomy-oriented response within the broader analysis of this supervision and evaluation bias.

\subsection*{Contributions}

The main contributions of this work are as follows:

\begin{itemize}

\item We formulate residual registration errors as structured geometric label noise in voxel-wise supervised synthetic CT generation, distinguishing their variable and systematic effects on the learned target distribution.

\item We empirically demonstrate that voxel-wise performance depends on the consistency between the registration conventions used to construct training targets and evaluation references, showing that standard metrics can reward agreement with the registration pipeline.

\item We analyze the consequences of registration-induced supervision bias for source-anatomy preservation, regression-induced blurring, OOD generalization, and predictive uncertainty.

\item We introduce and evaluate SAM-based perceptual supervision and a separately calibrated SAM-based evaluation metric as complementary anatomy-oriented tools beyond direct voxel-wise agreement.

\end{itemize}

These contributions are evaluated on 1,784 paired patients from six clinical centers, covering MR$\rightarrow$CT and CBCT$\rightarrow$CT synthesis across five anatomical regions, for a total of 196,845 image slices.

\section{Background}

\subsection{Image-to-Image Translation: Problem Formulation}
\label{sec:i2i_background}

Image-to-image translation transforms an image from a source domain into a target-domain representation while preserving the task-relevant content of the input \cite{mcnaughton2023machine,kaji2019overview}. In synthetic CT generation, this corresponds to predicting CT-like images from MRI or CBCT acquisitions for applications such as dose calculation, segmentation, and image registration \cite{roh2024ct,liu2021ct,dowling2022image,altalib2025synthetic}. This study focuses on supervised conditional synthesis, in which paired images provide voxel-level targets for model training.

\paragraph{Supervised translation as direct regression}
When spatially aligned pairs are available, synthesis models are commonly optimized using direct reconstruction losses such as the mean absolute error (MAE) or mean squared error (MSE).

From a statistical perspective, such objectives correspond to empirical risk minimization under pointwise loss functions. Let $x \in X$ denote an input image and $Y \sim p(y \mid x)$ the corresponding target random variable. Under a quadratic loss, the optimal predictor is the conditional expectation:
\begin{equation}
f^*(x) = \mathbb{E}[Y \mid X = x],
\end{equation}
whereas an $\ell_1$ loss yields the conditional median \cite{bishop2006pattern,koenker2001quantile}.

In image synthesis, voxel-wise regression losses are well known to produce oversmoothed predictions. This effect arises because pointwise objectives collapse the variability of the conditional distribution $p(y \mid x)$ into a single central-tendency estimate. Under common modeling assumptions, such objectives can also be interpreted as maximum-likelihood estimators under simple pixel-wise residual models, corresponding to independent Gaussian variability for $\ell_2$ objectives and Laplacian variability for $\ell_1$ objectives. Limited model capacity may also attenuate high-frequency details, but this approximation effect is conceptually distinct from uncertainty contained in the supervision targets.

\paragraph{Intrinsic modality ambiguity}
Cross-modality relationships can exhibit intrinsic ambiguity due to non-bijective signal formation, such as modality-specific contrast, partial volume effects, or artifacts, which induces a non-degenerate conditional distribution $p(y \mid x)$. In the settings considered here, anatomical correspondence is locally well constrained under ideal alignment, so this intrinsic variability is expected to remain limited and voxel-wise regression would not by itself induce severe blurring~\cite{rassmann2026regression}.

\paragraph{Registration-induced supervision uncertainty}
Voxel-wise supervision assumes that the input and target images are spatially aligned. In practice, the available training target is a registered reference $\tilde{y}$ that may differ from the ideal target $y$ because of residual registration errors:
\begin{equation}
    \tilde{y} = y \circ \phi,
\end{equation}
where $\circ$ denotes the spatial resampling operator and $\phi$ denotes the residual deformation field induced by imperfect registration. Training is consequently performed on the observed distribution $p(\tilde{y}\mid x)$ rather than on the ideal distribution $p(y\mid x)$.

Registration residuals do not behave as independent voxel-wise intensity noise. They produce spatially coherent displacements of anatomical structures and thus introduce structured geometric label noise into the training targets. Moreover, the residual deformation field $\phi$ is not necessarily purely random: it can contain pair-specific variability as well as systematic contributions determined by the similarity metric, regularization strategy, deformation model, or acquisition protocol.

These residuals affect $p(\tilde{y}\mid x)$ in two complementary ways. Their variable component increases the dispersion of the observed target distribution because similar local input configurations may be associated with target structures displaced differently across training pairs. Voxel-wise regression, which estimates a central tendency of this distribution, may consequently produce spatially averaged or blurred boundaries. Their systematic component can instead shift the center of the distribution when the registration pipeline consistently favors a particular geometric convention. The synthesis model may then learn this convention and reproduce the spatial behavior of the registration method used to construct the training data.

Registration-induced target uncertainty therefore provides an additional geometric explanation for the blurring observed in supervised synthesis, beyond intrinsic modality ambiguity. It cannot be resolved simply by increasing model capacity because the uncertainty lies in the training targets themselves. This setting was formalized as supervised learning with noisy labels by Kong et al.~\cite{kong2021breaking}, who proposed a joint registration-synthesis strategy that can theoretically recover the optimal clean-data solution under regularity assumptions on $\phi$. Their analysis, however, focuses on loss correction rather than on how uncorrected training reshapes $p(\tilde{y}\mid x)$ and affects uncertainty estimation and metric interpretation.

This form of uncertainty is distinct from both aleatoric and epistemic uncertainty. It is not inherent to the cross-modality mapping and does not primarily reflect limited model knowledge; instead, it originates from the data-construction pipeline. This distinction is rarely made explicit in the supervised synthesis literature. Blurring under MAE or MSE is commonly attributed to regression over an intrinsically ambiguous conditional distribution, whereas part of the observed ambiguity may arise from structured geometric inconsistencies in the registered targets.

\paragraph{Alleviating oversmoothing through perceptual and generative objectives}
One strategy for reducing oversmoothing is to augment voxel-wise reconstruction losses with a perceptual term computed in the feature space of a pretrained network \cite{johnson2016perceptual}. In its original formulation, the feature extractor is typically a VGG network trained on large-scale natural-image classification. The perceptual loss can be written as:
\begin{equation}
\begin{aligned}
    \mathcal{L}_{\mathrm{perceptual}}
    &= \|\psi(\hat{y}) - \psi(y)\|_1 ,
\end{aligned}
\end{equation}
where $\psi$ denotes the feature extractor and $\hat{y}=G_\theta(x)$ the synthesized image. As an $\ell_1$ objective in feature space, this loss can be interpreted as estimating a conditional median with respect to the representation induced by $\psi$, rather than directly in the voxel domain. By comparing representations that encode contextual and multi-scale structure, it can promote spatially coherent anatomical patterns and reduce the oversmoothing associated with voxel-wise regression.

The behavior of a perceptual loss nevertheless depends critically on the selected feature representation. Features learned from natural-image classification may not optimally encode the structures relevant to medical imaging. In this study, SAM embeddings are used because they provide multi-scale representations of anatomical structures and boundaries. The resulting distance is intended to complement voxel-wise metrics by being less sensitive to small residual displacements while remaining responsive to meaningful structural differences, as detailed in Section~\ref{sec:sam_supervision}.

Distribution-level objectives, including adversarial and diffusion-based approaches, can similarly improve sharpness and visual realism by counteracting the averaging behavior of voxel-wise regression \cite{isola2017image,ho2020denoising}. However, perceptual and generative objectives do not remove the systematic component of registration-induced uncertainty when they are trained on the same imperfectly aligned pairs. They still learn from $p(\tilde{y}\mid x)$ rather than from $p(y\mid x)$ and may therefore reproduce biases embedded in the registration pipeline.

\paragraph{Consequences for synthetic CT evaluation}
Voxel-wise supervised translation is well suited to settings with reliable spatial correspondence. When residual registration errors are structured or systematic, however, regression objectives may both smooth uncertain boundaries and shift anatomical structures toward the geometric convention imposed by the registration pipeline.

This limitation remains insufficiently characterized in the recent literature \cite{florkow2019impact,zimmermann2026eliminating,rossi2021comparison}, where evaluation protocols predominantly emphasize intensity-based similarity rather than anatomical faithfulness. Benchmarking may consequently favor models that reproduce registration-induced distortions rather than models that best preserve clinically relevant structures. This is particularly problematic when the clinical motivation for synthetic CT is to avoid uncertain inter-modality registration, for example by generating CT-like images directly from MRI.

Improving spatial correspondence is a natural response, and several methods jointly optimize synthesis and registration \cite{kong2021breaking,li2025boosting,xin2024deformation}. Nevertheless, residual misalignment remains difficult to eliminate completely, especially under large anatomical deformations. Unsupervised or weakly paired methods relax the requirement for exact voxel-wise correspondence, but introduce other challenges related to anatomical consistency, training stability, and quantitative validation. The present study therefore focuses on quantifying registration-induced bias in supervised synthesis and examining its consequences for uncertainty estimation and sCT evaluation, as further discussed in Section~\ref{sec:discussion_bias}.

\section{Supervised Cross-Modality Image Synthesis: Experimental Study on SynthRAD}

In this section, we consider the standard supervised formulation of cross-modality image synthesis, as commonly adopted in recent benchmarks. Given spatially aligned image pairs, a model is trained to predict CT images from input CBCT or MRI data using voxel-wise reconstruction losses.

This setting constitutes the dominant paradigm in current evaluation frameworks, where performance is primarily assessed through intensity-based similarity metrics computed with respect to a reference CT.

\subsection{SynthRAD challenge and datasets}

Experiments were conducted using the SynthRAD2023 and SynthRAD2025 datasets \cite{thummerer2025synthrad2025,huijben2024generating}, which contain paired images for two synthesis tasks: MR-to-CT (Task 1) and CBCT-to-CT (Task 2). The datasets cover brain, head-and-neck, thoracic, abdominal, and pelvic anatomies, depending on the task and challenge edition. Training pairs were provided after rigid registration, whereas the hidden validation and test sets enabled evaluation under the official challenge protocol. This structure makes the dataset particularly suitable for studying how the choice of registration convention affects both model training and performance assessment.

The official image-similarity metrics were mean absolute error (MAE), peak signal-to-noise ratio (PSNR), and multi-scale structural similarity (MS-SSIM); the local analyses report SSIM. Dose-based metrics were also reported by the challenge but were
used here only to document the official benchmark performance. Dataset composition and complete evaluation details are provided in Appendix~\ref{app:synthrad} and in the SynthRAD references \cite{thummerer2025synthrad2025,huijben2024generating}.

\subsection{Experimental setup}

\begin{figure*}[h!]
  \centering
  \includegraphics[width=0.95\textwidth,height=0.75\textheight,keepaspectratio]{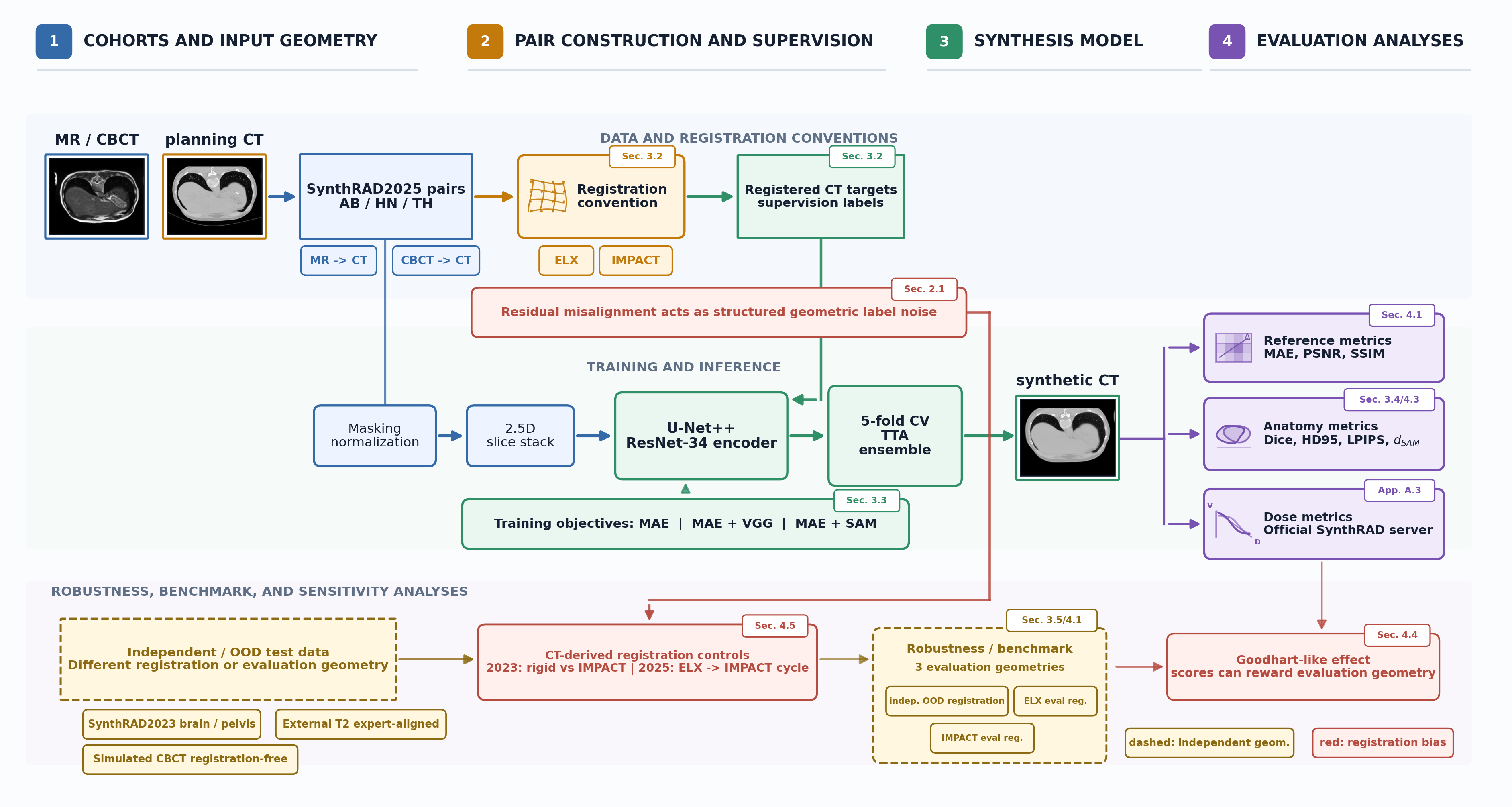}
    \caption{Overview of the synthesis and evaluation protocol used to study registration-induced bias. Paired MR/CBCT and planning CT data are first converted into supervised training pairs through ELX or IMPACT registration conventions, yielding registered CT targets used as supervision labels for a 2.5D U-Net++ trained with MAE, VGG, or SAM-based objectives. Rectangular blocks denote data, targets, or outputs, whereas rounded blocks denote processing modules, training components, metrics, or analysis toolboxes; dashed blocks indicate independent or OOD evaluation geometries. The resulting synthetic CT images are evaluated using reference-based, anatomy-oriented, and dose-based metrics, complemented by independent/OOD test settings and CT-derived controls that isolate robustness and registration-induced metric sensitivity. Small tags indicate where the corresponding protocol components and analyses are defined in the manuscript. The robustness and benchmark branch distinguishes the independent/OOD geometry from ELX- and IMPACT-based evaluation registrations.}
\label{fig:architecture}
\end{figure*}

The overall study design is summarized in Fig.~\ref{fig:architecture}, which links the construction of registered supervision targets, the supervised synthesis model, and the complementary evaluation analyses used to assess registration-induced bias.

\paragraph{Data and registration conventions}

Although the dataset is provided as paired multimodal acquisitions, the correspondence between modalities is not intrinsically voxel-wise. MRI, CBCT, and CT scans are acquired at different time points and under different physiological conditions, leading to non-negligible anatomical discrepancies.

To study the impact of spatial correspondence on supervised synthesis while remaining consistent with the official challenge evaluation protocol, we investigate two different registration strategies for constructing the training pairs.

First, we consider the registration pipeline used for the test set by the challenge organizers. This approach relies on a multi-resolution deformable registration framework implemented in Elastix \cite{klein2009elastix}, using a mutual-information-based similarity metric representative of standard practice in radiotherapy workflows and supervised sCT synthesis studies. 
In the remainder of this manuscript, this registration configuration is referred to as \textbf{ELX}.

Second, we investigate an alternative registration strategy based on IMPACT-Reg \cite{boussot2025impact}, a feature-based multimodal registration method leveraging deep semantic representations. 
Unlike intensity-based approaches, IMPACT-Reg relies on high-level feature correspondences to estimate multimodal anatomical alignment.
In the remainder of this manuscript, this registration configuration is referred to as \textbf{IMPACT}. 
Neither ELX nor IMPACT is assumed to provide ground-truth anatomical correspondence. They are treated as two alternative registration conventions whose residual errors may differ. IMPACT is used here because its alignments show higher anatomical consistency than ELX according to the segmentation-based analyses reported below, not because it is considered an exact reference.

\subsubsection{Training and inference}

All images were resampled and intensity-normalized before training. We used a 2.5D U-Net++ architecture with a ResNet-34 encoder \cite{zhou2018unet++} that predicts the central sCT slice from adjacent input slices. Models were optimized with the loss described below, and checkpoints were selected according to validation MAE. Predictions from the retained models were averaged at inference time. Complete preprocessing, architecture, optimization, and inference parameters are provided in Appendix~\ref{app:implementation}.

The objective of this work is not to introduce a new architecture, but to use a reliable and efficient supervised baseline for studying the effect of registration-induced supervision bias. The 2.5D U-Net++ provides a practical compromise between anatomical context, computational cost, and deployment simplicity.

\subsection{SAM-based perceptual supervision}
\label{sec:sam_supervision}

We investigate the impact of SAM-based perceptual supervision by comparing two training objectives: a standard voxel-wise reconstruction loss and an augmented objective combining voxel-wise and feature-based constraints. This subsection describes only the losses used for model optimization; the separately calibrated SAM-based evaluation metric is introduced in Section~\ref{sec:sam_metric}.

\paragraph{Voxel-wise reconstruction loss}
The baseline model is trained using a standard $\ell_1$ loss, defined as:
\begin{equation}
\mathcal{L}_{\text{MAE}} = \left\| \hat{y} - y \right\|_1,
\end{equation}
where $\hat{y}$ denotes the synthesized CT image and $y$ the reference CT.

This formulation corresponds to the dominant paradigm in supervised sCT generation, where the model is optimized to minimize voxel-wise discrepancies between paired images, in accordance with the MAE-based evaluation criterion commonly used in the field.

\paragraph{SAM-based perceptual loss}
To complement voxel-level supervision, we introduce a perceptual loss defined in the feature space of a pretrained segmentation model.

Classical perceptual losses commonly rely on networks trained on natural images (e.g., VGG). Instead, we use the frozen encoder of the Segment Anything Model (SAM), in its SAM~2 version based on the Hiera hierarchical transformer architecture, as a feature extractor \cite{ravi2024sam2}. The encoder parameters remain fixed throughout synthesis-model training and provide multi-scale feature representations for comparing the synthesized and reference CT images.

Let $\psi(\cdot)$ denote the frozen SAM encoder. The perceptual loss is defined as:
\begin{equation}
\begin{aligned}
\mathcal{L}_{\text{SAM}}
&= \sum_{l \in \mathcal{S}} w_l\,\Delta_l, \\
\Delta_l
&= \left\| \psi_l(\hat{y}) - \psi_l(y) \right\|_1 ,
\end{aligned}
\end{equation}
where $\psi_l(\cdot)$ denotes the feature map extracted at layer $l$.

In practice, features are extracted from four hierarchical levels of the encoder. 
Only the two intermediate feature maps are used, corresponding to a weighting scheme of $(0, 1, 1, 0)$. 
This selection targets representations that capture anatomical structures at an appropriate level of abstraction, avoiding both low-level noise sensitivity and overly coarse semantic features.

\paragraph{Combined objective}
The final training objective is defined as:
\begin{equation}
\mathcal{L}_{\text{total}} = \mathcal{L}_{\text{MAE}} + \mathcal{L}_{\text{SAM}},
\end{equation}
where both terms are equally weighted.

Rather than relying on extensive hyperparameter tuning, this formulation is intentionally kept simple to assess the intrinsic contribution of perceptual supervision. 
A method that remains effective under minimal tuning is more likely to generalize across datasets and clinical conditions.

\subsection{SAM-based perceptual evaluation metric}
\label{sec:sam_metric}

Separately from the training objective described in Section~\ref{sec:sam_supervision}, we define a calibrated SAM-based perceptual distance for anatomy-oriented evaluation. The SAM-based metric is computed only after model training and is not used to optimize the synthesis network or to select its checkpoints. In the spirit of LPIPS \cite{zhang2018unreasonable}, SAM provides the feature representation, while separately learned channel-wise weights calibrate the contribution of the selected features to the final distance.

The objective is to calibrate the feature-space distance so that it better discriminates the structural defects typically produced by MAE-trained synthetic CT models from the residual discrepancies caused by imperfect registration.

Starting from the SAM-based distance defined above, we introduce channel-wise weights $\alpha_{l,c}$ applied to the feature discrepancies at each selected layer. These weights are optimized from precomputed feature differences rather than fixed heuristically. The calibration is designed to emphasize feature channels that distinguish a sharp CT image affected by residual alignment errors from a synthetic CT prediction affected by regression-induced blurring or structural inaccuracies.

Calibration is performed once on development cases only, before final evaluation. For each selected SAM layer, absolute feature differences are spatially averaged per channel and normalized using calibration-set statistics. Non-negative channel weights are then learned with a hinge-ranking objective:
\begin{equation}
\begin{aligned}
\mathcal{L}_{\mathrm{cal}}
&= \frac{1}{N}\sum_i
\max\bigl(0, m + d_i^{\mathrm{def}} - d_i^{\mathrm{MAE}}\bigr)
+ \lambda\|\alpha\|_2^2, \\
d_i^{\mathrm{def}}
&= d_{\alpha}(\mathrm{CT}_{\mathrm{def}}^i,\mathrm{CT}^i), \\
d_i^{\mathrm{MAE}}
&= d_{\alpha}(\mathrm{sCT}_{\mathrm{MAE}}^i,\mathrm{CT}^i),
\end{aligned}
\end{equation}
where $\mathrm{CT}_{\mathrm{def}}$ denotes the deformed CT control, $m$ is a fixed margin, $\lambda$ controls a small $\ell_2$ regularization term, and the learned weights are normalized after optimization. The final weights are frozen and reused for all reported evaluations. This constraint encodes the working assumption that a deformed CT control, although still affected by residual registration errors, preserves CT-like structural detail and high-frequency anatomy better than a purely MAE-trained synthetic prediction.

The resulting calibrated metric downweights feature responses dominated by residual misalignment and emphasizes channels sensitive to the characteristic defects of MAE-trained synthesis, such as blurring, loss of fine anatomical boundaries, or structurally inconsistent predictions. The optimization identifies the intermediate SAM feature levels, particularly layers 2 and 3, as the most discriminative representations. This suggests that early features are too local and sensitive to low-level appearance differences. In contrast to an unweighted perceptual distance, the optimized SAM-based metric is therefore designed to better reflect structural defects in synthetic CT images.
Accordingly, $\mathcal{L}_{\mathrm{SAM}}$ denotes the feature-space loss used during synthesis-model training, whereas $d_{\mathrm{SAM}}$ denotes the separately calibrated distance used for evaluation.

\subsection{Experimental configurations}
In the experimental study, we compare two synthesis objectives and two registration strategies:
\begin{itemize}
\item \textbf{MAE:} synthesis model trained with the voxel-wise reconstruction loss $\mathcal{L}_{\text{MAE}}$ only;
\item \textbf{SAM:} synthesis model trained with the combined objective $\mathcal{L}_{\text{total}}$, including SAM-based perceptual supervision;
\item \textbf{ELX:} training pairs constructed using the official Elastix-based registration pipeline;
\item \textbf{IMPACT:} training pairs constructed using the IMPACT-Reg registration pipeline.
\end{itemize}

\paragraph{Evaluation protocol}

We adopt a held-out evaluation protocol combined with five-fold cross-validation for model development. For each task, 15\% of the available cases are excluded from training and reserved for final testing, while the remaining data are used to train five independent models subsequently combined through ensembling at inference time. The evaluation branches in Fig.~\ref{fig:architecture} summarize the three complementary analyses used below: reference-based metrics, anatomy- and dose-oriented assessment, and robustness/sensitivity controls.

\paragraph{Task 1 (MRI-to-CT)}

The final evaluation protocol relies on three complementary test settings.

\textbf{(1) In-distribution (ID) evaluation.} 
The ID test set corresponds to the 66 held-out cases from SynthRAD2025 (AB = 22, HN = 21, TH = 23). 
For this subset, two versions of the data are considered: the ELX-based non-rigid registration provided by the organizers and our IMPACT-based non-rigid registration. These two aligned references are used to evaluate the sensitivity of quantitative metrics to the choice of registration under in-distribution conditions.

\textbf{(2) Out-of-distribution (OOD) evaluation with SynthRAD2023.} 
The OOD test set includes 50 cases from SynthRAD2023 (brain = 27, pelvis = 23). 
Similarly to the ID setting, both ELX-based and IMPACT-based registrations are used to generate two evaluation references. Notably, the ELX registration corresponds to a rigid alignment, consistent with the original evaluation setup of the challenge, while IMPACT provides a non-rigid alternative. This setting allows us to assess the impact of registration differences in a more challenging OOD scenario.

\textbf{(3) OOD evaluation with expert-refined alignment (Ext-T2).} 
Finally, we evaluate the model on an external OOD dataset composed of 24 T2-weighted MRI cases \cite{dowling2015automatic}. 
This dataset represents a particularly severe OOD setting, as no T2-weighted MRI data are included in the training set. 
The dataset further provides high-quality voxel-wise alignment between MRI and CT, obtained through a single reference alignment manually refined by experts. 

In contrast to the previous settings, this dataset relies on a unique reference, removing variability induced by different registration methods and providing an expert-refined evaluation setting with reduced registration uncertainty.

\paragraph{Task 2 (CBCT-to-CT)}

The evaluation protocol for Task 2 relies on two complementary test settings based on the SynthRAD2025 dataset.

\textbf{(1) Real CBCT evaluation.} 
The first setting uses the 103 held-out SynthRAD2025 cases, where CBCT and CT are acquired independently (AB = 32, HN = 37, TH = 34). 
Similarly to Task 1, two versions of the data are considered using ELX-based and IMPACT-based registrations to define the evaluation reference. 

\textbf{(2) Simulated CBCT evaluation (Sim-CBCT, registration-free).} 
To isolate the effect of registration bias, we construct a second evaluation set by generating simulated CBCT volumes from the same 103 CT images. 
These simulated CBCT volumes are obtained by forward-projecting the planning CT using an RTK-based \cite{rit2014reconstruction} cone-beam simulation pipeline, followed by projection degradation and FDK reconstruction.

Because the simulated CBCT is generated directly from the CT volume, the anatomical correspondence between input and reference is preserved by construction. This removes the inter-modality registration uncertainty present in real CBCT-CT pairs and provides a controlled setting where quantitative metrics directly reflect synthesis quality.

However, the simulated CBCT volumes do not perfectly reproduce real acquisitions. Hounsfield units are not fully calibrated, and some CT volumes are truncated, leading to differences in field-of-view compared to clinical CBCT scans. The simulation therefore does not strictly match the physical acquisition process of real CBCT, and noticeable intensity and boundary discrepancies may arise. Nevertheless, the simulated data capture the main characteristics of CBCT imaging, providing a realistic yet controlled approximation suitable for analysis.

In the result tables, AB/HN/TH denotes the real held-out CBCT setting, whereas $AB_{\mathrm{sim}}$, $HN_{\mathrm{sim}}$, and $TH_{\mathrm{sim}}$ denote these registration-free simulated CBCT evaluations, and Sim-CBCT their aggregate. For Task 1, Ext-T2 denotes the external T2-weighted MRI set.

\subsection{Statistical analysis}
\label{sec:statistics}

All quantitative metrics were first computed at the patient-volume level. For each anatomical region and experimental configuration, results are reported as the mean and standard deviation across patients. Comparisons between models or registration configurations were performed on matched patients using two-sided Wilcoxon signed-rank tests. The patient, rather than the slice or the cross-validation fold, was treated as the statistical unit. Statistical significance is denoted by $^{*}p_{\mathrm{adj}}<0.05$, $^{**}p_{\mathrm{adj}}<0.01$, and $^{***}p_{\mathrm{adj}}<0.001$.

For ensemble results (CV), the five cross-validation predictions were averaged voxel-wise for each patient before computing the evaluation metrics. For Mean CV results, each metric was first computed for every fold-specific prediction and patient, then averaged across the five folds for that patient before calculating group-level summary statistics. Thus, folds were not treated as independent statistical observations.

\subsection{Qualitative comparison of IMPACT and ELX registrations}

As a preliminary step before assessing the impact of registration on training and evaluation in cross-modality image synthesis, qualitative comparisons between IMPACT- and ELX-based registrations across several anatomical regions (AB, HN, TH) in the MR$\rightarrow$CT setting are provided in Appendix Figure~\ref{fig:sct_mr_comparison}.

Both registration approaches produce visually plausible and globally coherent alignments across the three anatomical regions considered. At the scale of the full volume, structural correspondences between MR and CT are largely established in both cases, although residual anatomical discrepancies remain visible in regions subject to large deformations, including the diaphragm and soft tissue boundaries. This stands in contrast to the rigid-only alignment adopted in the SynthRAD2023 edition, which left substantially larger residual discrepancies between modalities. In the present setting, both ELX and IMPACT perform deformable non-rigid registration, and ELX already represents the level of registration quality commonly used in supervised synthetic CT studies. The observed differences should therefore be interpreted as refinements over an already credible and clinically realistic alignment rather than as a fundamental reliability gap. It is also important to note that the effects reported in the following sections are obtained despite this relatively strong registration baseline; under the rigid-only alignment setup used in the SynthRAD2023 edition, the observed differences and their impact on supervised synthesis would likely have been substantially larger.

These qualitative observations are complemented by the structure-wise Dice analysis reported in Table~\ref{tab:diceimpactreg}. The table summarizes the mean Dice coefficient obtained after registration for each anatomical region. Higher Dice values indicate greater agreement between the registered CT structures and the source-modality anatomy captured by the segmentation analysis. They therefore support the use of IMPACT as an alternative supervision convention with higher measured anatomical consistency on average, without establishing it as a ground-truth alignment.

\begin{table}[h!]
\centering
\caption{Mean Dice coefficient per anatomical region for the evaluated registration approaches. Higher values indicate better anatomical alignment between the registered source-modality structures and the reference CT structures.}
\label{tab:diceimpactreg}
\begin{tabular}{lccc}
\toprule
\textbf{Region} & \textbf{Rigid} & \textbf{ELX} & \textbf{IMPACT} \\
\midrule
Pelvis & 0.64 & 0.70 & \textbf{0.73} \\
Abdomen & 0.63 & \textbf{0.71} & 0.70 \\
Thorax & 0.61 & 0.66 & \textbf{0.71} \\
Head \& Neck & 0.72 & 0.75 & \textbf{0.81} \\
Brain & 0.71 & 0.80 & \textbf{0.84} \\
\midrule
\textbf{Mean} & 0.662 & 0.724 & \textbf{0.758} \\
\bottomrule
\end{tabular}

\end{table}

Overall, IMPACT achieves a modest average Dice increase over ELX, from 0.724 to 0.758, although the direction of the difference is not uniform across every region. Together with the qualitative observations, these results define two alternative supervision conventions with different measured levels of anatomical consistency. 

\section{Results}

\subsection{Effect of Registration Consistency Between Training and Evaluation}

Table~\ref{tab:registration_consistency_aggregate} summarizes the aggregate in-distribution results for all combinations of registration strategies used during training and evaluation. The first term in each configuration denotes the registration used to construct the training targets, whereas the second denotes the registration used to define the evaluation reference. Complete region-wise results, including patient-level variability and statistical comparisons, are reported in Appendix~\ref{app:registration_results}, Tables~\ref{tab:task1_cv_region_metrics_in_distribution} and~\ref{tab:task2_cv_region_metrics_in_distribution}.

\begin{table}[h!]
\centering
\fontsize{7.5}{9}\selectfont
\setlength{\tabcolsep}{2.2pt}
\renewcommand{\arraystretch}{1.18}
\caption{Aggregate in-distribution performance for all combinations of registration strategies used during training and evaluation. Results are pooled across the AB, HN, and TH regions. Complete regional results are reported in Appendix~\ref{app:registration_results}.
}
\label{tab:registration_consistency_aggregate}
\begin{tabular}{@{}lcccccc@{}}
\toprule
& \multicolumn{3}{c}{Task 1} & \multicolumn{3}{c}{Task 2} \\
\cmidrule(lr){2-4} \cmidrule(lr){5-7}
Train/Eval
& MAE $\downarrow$ & PSNR $\uparrow$ & SSIM $\uparrow$
& MAE $\downarrow$ & PSNR $\uparrow$ & SSIM $\uparrow$ \\
\midrule
IMPACT/IMPACT
& \textbf{63.55} & \textbf{29.97} & \textbf{0.927}
& \textbf{58.76} & \textbf{31.16} & \textbf{0.937} \\
IMPACT/ELX
& 72.47 & 28.49 & 0.918
& 69.17 & 29.13 & 0.920 \\
ELX/IMPACT
& 68.31 & 29.37 & 0.920
& 65.70 & 30.08 & 0.921 \\
ELX/ELX
& 66.98 & 29.24 & 0.924
& 61.28 & 30.22 & 0.930 \\
\bottomrule
\end{tabular}

\end{table}
\renewcommand{\arraystretch}{1}

Across both tasks, the best voxel-wise performance is obtained when the registration strategy used for training and evaluation is matched. In Task 1, IMPACT/IMPACT reaches an aggregate MAE of 63.55, compared with 72.47 when the same model is evaluated against ELX references. In Task 2, the same pattern is stronger, with MAE increasing from 58.76 to 69.17 under evaluation mismatch. The effect is asymmetric: evaluation mismatch degrades IMPACT-trained models by 14.0\% (Task~1) and 17.7\% (Task~2), but ELX-trained models by only 2.0\% and 7.2\%, which still remain above IMPACT/IMPACT. This behavior indicates that supervised models do not only learn a modality translation mapping, but also partially adapt to the geometric convention imposed by the registration pipeline.

This effect becomes more pronounced in OOD regions. The same dependence is observed in the SynthRAD2023 brain and pelvis regions, where the difference between the rigid-only ELX convention and the IMPACT deformable convention is larger. In these regions, ELX-trained models even score better against IMPACT than against the rigid ELX references, consistent with their deformable training targets being geometrically closer to IMPACT than to a rigid alignment. IMPACT/IMPACT nevertheless remains the best configuration. Full regional results are provided in Appendix~\ref{app:registration_results}.

Interestingly, the top-performing methods reported in the original SynthRAD2023 challenge achieved substantially lower voxel-wise errors under this rigid-registration setting, reaching $58.83 \pm 13.41$ HU MAE, $29.61 \pm 1.79$ dB PSNR, and $0.885 \pm 0.029$ SSIM for Task~1. These results were obtained using a supervised synthesis framework conceptually close to the one proposed in this work, but trained directly on rigidly aligned image pairs, therefore matching the registration convention used during evaluation. This further indicates that strong in-distribution voxel-wise performance can still be achieved even with imperfectly aligned training pairs, provided that the same alignment convention is consistently used during both training and evaluation.

Table~\ref{tab:ood_combined} reports performance in OOD settings where the evaluation references differ from the two registration methods used during training (ELX and IMPACT). IMPACT-trained models consistently achieve lower MAE and higher PSNR and SSIM than ELX-trained models across both tasks, with all differences reaching $p<0.001$. The MAE improvement ranges from 9.0\% to 15.0\% across the Task 2 simulated CBCT regions, supporting the conclusion that the effect is not limited to agreement with the IMPACT evaluation reference.

\begin{table}[h!]
\centering
\scriptsize
\setlength{\tabcolsep}{4pt}
\caption{
Quantitative results on out-of-distribution (OOD) regions for Task 1 and Task 2 in supervised cross-validation experiments. Mean $\pm$ standard deviation of commonly used image similarity metrics (MAE, PSNR, SSIM) are reported. Column groups indicate the models used (IMPACT or ELX). Ext-T2 is the external T2-weighted MRI set; $AB_{\mathrm{sim}}$, $HN_{\mathrm{sim}}$, and $TH_{\mathrm{sim}}$ are the registration-free simulated CBCT sets. Statistical comparisons against IMPACT follow the patient-level protocol described in Section~\ref{sec:statistics}.
}
\label{tab:ood_combined}
\begin{tabular}{llcccccc}
\toprule
Task & Region & \multicolumn{3}{c}{IMPACT} & \multicolumn{3}{c}{ELX} \\
\cmidrule(lr){3-5} \cmidrule(lr){6-8}
 &  & MAE & PSNR & SSIM & MAE & PSNR & SSIM \\
\midrule

\multirow{1}{*}{T1} 
& Ext-T2
& \bmstd{88.54}{12.51}{} 
& \bmstd{26.82}{0.91}{} 
& \bmstd{0.899}{0.025}{} 
& \mstd{97.43}{14.83}{***} 
& \mstd{26.19}{0.87}{***} 
& \mstd{0.889}{0.031}{***} \\

\midrule

\multirow{3}{*}{T2}
& $AB_{\mathrm{sim}}$
& \bmstd{60.41}{14.83}{} 
& \bmstd{30.32}{1.87}{} 
& \bmstd{0.927}{0.016}{} 
& \mstd{71.09}{13.12}{***} 
& \mstd{28.68}{1.30}{***} 
& \mstd{0.915}{0.014}{***} \\

& $HN_{\mathrm{sim}}$
& \bmstd{136.18}{30.01}{} 
& \bmstd{24.10}{2.05}{} 
& \bmstd{0.901}{0.029}{} 
& \mstd{149.66}{28.91}{***} 
& \mstd{23.45}{1.65}{***} 
& \mstd{0.892}{0.031}{***} \\

& $TH_{\mathrm{sim}}$
& \bmstd{95.99}{44.86}{} 
& \bmstd{27.38}{3.26}{} 
& \bmstd{0.885}{0.062}{} 
& \mstd{109.15}{49.65}{***} 
& \mstd{26.34}{2.93}{***} 
& \mstd{0.870}{0.067}{***} \\

\bottomrule
\end{tabular}

\end{table}

\begin{figure*}[h!]
  \centering
  \includegraphics[width=0.96\textwidth,height=0.36\textheight,keepaspectratio]{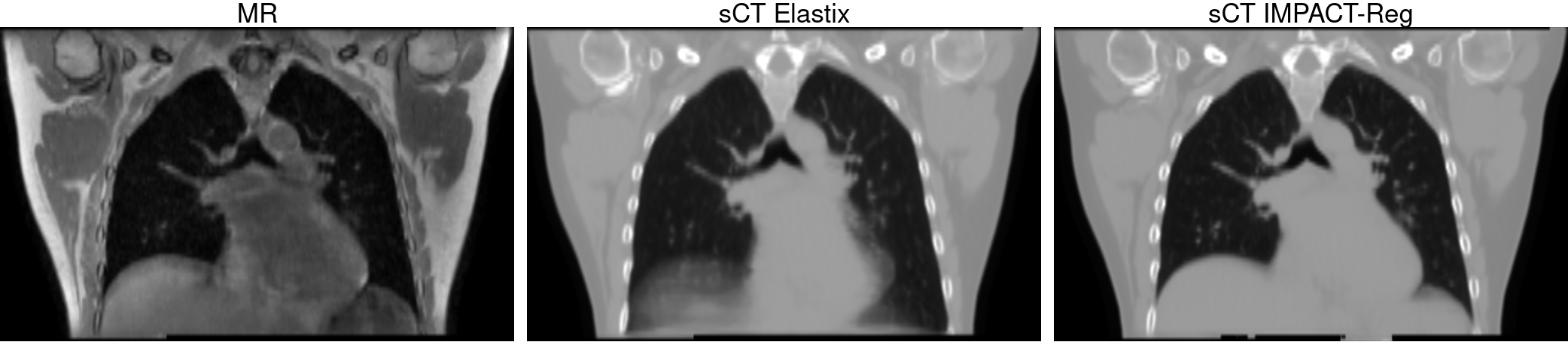}
\caption{Qualitative comparison between MR input and synthesized CT generated using ELX-based and IMPACT-based training.}
\label{fig:sct_mr_comparison_synthesis}
\end{figure*}

Figure~\ref{fig:sct_mr_comparison_synthesis} presents a qualitative comparison of synthetic CT images generated from the same MR input using ELX- and IMPACT-based training. Both models produce visually plausible CT-like outputs with globally coherent intensity distributions. However, a clear anatomical inconsistency is observed in the ELX-trained output: the diaphragm region is severely distorted, with the cardiac silhouette and inferior lung boundaries displaced in a manner that does not correspond to the source MR anatomy. This deformation is not present in the MR input and represents a spurious anatomical configuration introduced by the synthesis model. In contrast, the IMPACT-trained model, trained using more geometrically consistent image pairs than ELX, preserves the diaphragm position and overall thoracic anatomy in a configuration consistent with the source image, with sharper pulmonary contours and more faithful mediastinal structure delineation.

Table~\ref{tab:cv_uncertainty_mae_percentage} presents prediction uncertainty estimated from the variance across 15 synthesized predictions per patient, corresponding to the combination of five cross-validation models and three test-time augmentations (original, horizontal flip, and vertical flip). Results are reported for models trained using the voxel-wise MAE criterion. Uncertainty is systematically higher for ELX-trained models than for IMPACT-trained models across both tasks. In Task~1, in-distribution regions show relative uncertainty increases ranging from 20.7\% (HN and TH) to 41.2\% (AB), with an aggregate increase of 29.7\%. In Task~2, the same trend is observed but with more variable magnitude, ranging from 1.2\% (HN) to 27.6\% (AB), with an aggregate increase of 12.5\%. In Task~1, uncertainty differences become substantially larger in OOD settings, reaching 116.1\% for the pelvis, while the Ext-T2 increase reaches 61.8\%. The brain region constitutes a partial exception, with a more limited uncertainty increase of 2.7\%, consistent with its lower overall performance gap between registration strategies. Overall, ELX-trained models exhibit greater prediction variability than IMPACT-trained models, indicating that synthesized predictions vary more strongly from one inference configuration to another when the models are trained on less geometrically consistent image pairs.

\begin{table}[h!]
\centering
\small
\setlength{\tabcolsep}{6pt}
\renewcommand{\arraystretch}{1.15}
\caption{
Relative percentage increase of uncertainty for ELX compared to IMPACT. 
Uncertainty is estimated as the variance across the 15 predictions per patient obtained from the five cross-validation models and three test-time augmentations. Ext-T2 denotes the external T2-weighted MRI set and Sim-CBCT the registration-free simulated CBCT set.
}
\label{tab:cv_uncertainty_mae_percentage}
\begin{tabular}{lcc}
\toprule
Region & Task 1 & Task 2 \\
\midrule
AB & +41.2\% & +27.6\% \\
HN & +20.7\% & +1.2\% \\
TH & +20.7\% & +12.1\% \\
AB/HN/TH & +29.7\% & +12.5\% \\
\addlinespace[0.25em]
\hline
\hline
\addlinespace[0.25em]
Brain & +2.7\% & -- \\
Pelvis & +116.1\% & --\\
Ext-T2 & +61.8\% & -- \\
Sim-CBCT & -- & +1.8\% \\
\bottomrule
\end{tabular}

\end{table}

\subsection{SAM-based perceptual supervision}

Table~\ref{tab:sam_dice_compact} compares MAE-, VGG-, and SAM-based supervision using downstream Dice scores obtained with
TotalSegmentator. Because this evaluation does not reuse SAM features, it provides an independent assessment of anatomical preservation. Complete Dice and SSIM results are reported in Appendix~\ref{app:sam_results}.

\begin{table}[h!]
\centering
\small
\setlength{\tabcolsep}{5pt}
\renewcommand{\arraystretch}{1.15}
\caption{Downstream Dice scores obtained after training with MAE, VGG-based perceptual, or SAM-based perceptual supervision. All models use IMPACT-based training pairs. For Task 2, $AB_{\mathrm{sim}}$, $HN_{\mathrm{sim}}$, and $TH_{\mathrm{sim}}$ denote registration-free simulated CBCT evaluations. Complete region-wise Dice and SSIM results are reported in Appendix~\ref{app:sam_results}.}
\label{tab:sam_dice_compact}
\begin{tabular}{llccc}
\toprule
Task & Evaluation set & MAE & VGG & SAM \\
\midrule
Task 1 & AB/HN/TH & 0.711 & 0.712 & \textbf{0.738} \\
Task 1 & Brain/Pelvis & 0.749 & 0.772 & \textbf{0.806} \\
Task 1 & Ext-T2 & 0.665 & 0.672 & \textbf{0.725} \\
\midrule
Task 2 & AB/HN/TH & 0.695 & 0.683 & \textbf{0.700} \\
Task 2 & $AB_{\mathrm{sim}}$ & 0.709 & 0.723 & \textbf{0.764} \\
Task 2 & $HN_{\mathrm{sim}}$ & 0.734 & 0.750 & \textbf{0.755} \\
Task 2 & $TH_{\mathrm{sim}}$ & 0.722 & 0.741 & \textbf{0.758} \\
\bottomrule
\end{tabular}

\end{table}
\renewcommand{\arraystretch}{1}

SAM-based supervision improves the aggregate Task 1 Dice from 0.711 with MAE and 0.712 with VGG to 0.738. The improvement is larger in
unseen regions, reaching 0.806 for brain/pelvis and 0.725 on Ext-T2. In Task 2, the aggregate in-distribution improvement
is smaller, but SAM obtains the best Dice in all three simulated CBCT regions. SSIM remains similar to, or occasionally lower than, that obtained with MAE supervision, indicating that improved anatomical preservation is not always reflected by reference-based intensity similarity.

\begin{figure*}[h!]
  \centering
  \includegraphics[width=0.84\textwidth,height=0.74\textheight,keepaspectratio]{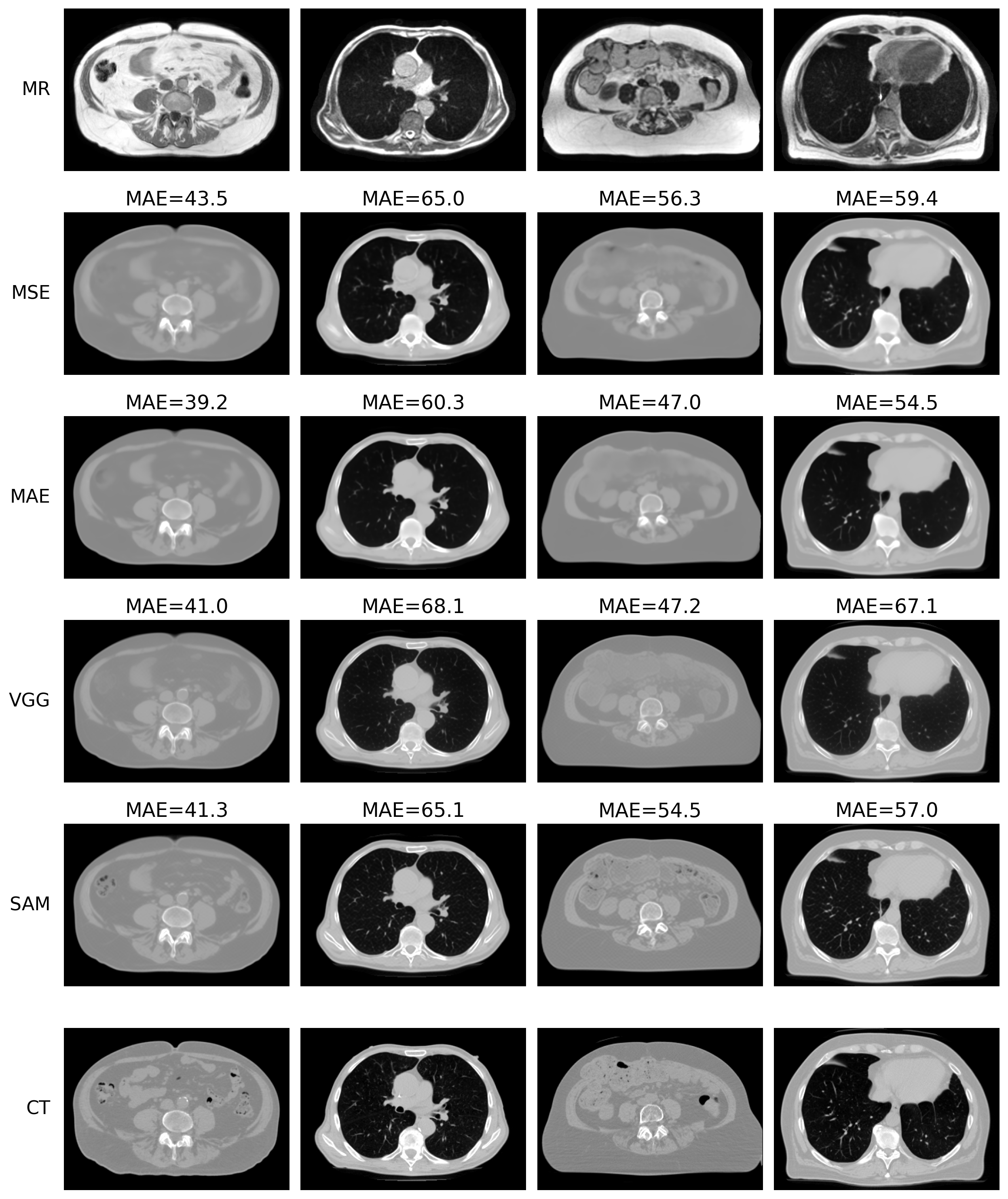}
  \caption{Qualitative comparison of synthesized CT images obtained using different training losses (rows) across multiple anatomical cases (columns). From top to bottom: input MR images, sCT generated using MSE, MAE, MAE+VGG-based perceptual loss, and MAE+SAM-based loss, followed by the reference CT images. For each synthesized image, the corresponding MAE with respect to the reference CT is reported. While all methods produce visually plausible outputs, differences in image sharpness and structural fidelity can be observed across losses, particularly at anatomical boundaries.}
  \label{fig:sam_comparison}
\end{figure*}

Figure~\ref{fig:sam_comparison} illustrates the same trend qualitatively. Voxel-wise objectives produce smoother predictions, whereas perceptual supervision, particularly with SAM features, better preserves sharp boundaries and fine anatomical structures. These visual differences are consistent with the downstream Dice improvements.

Table~\ref{tab:sam_perceptual_compact} reports both the average performance of the individual cross-validation models (Mean CV) and the performance of their voxel-wise ensemble (CV ensemble). Complete Mean CV and region-wise results are reported in Appendix~\ref{app:perceptual_results}, Tables~\ref{tab:task1_elx_fold_vs_cv_mae_median_perceptual} and~\ref{tab:task2_elx_fold_vs_cv_mae_median_perceptual}.
In the in-distribution regions, SAM supervision consistently improves $d_{\mathrm{SAM}}$ and LPIPS but increases MAE relative to direct MAE supervision. This trade-off is observed for both individual models and their ensembles, suggesting that it reflects the training objective rather than ensemble construction.

\begin{table*}[h!]
\centering
\scriptsize
\setlength{\tabcolsep}{4pt}
\renewcommand{\arraystretch}{1.15}
\caption{Aggregate performance of models trained with MAE or SAM-based supervision and evaluated with ELX references. Mean CV denotes the patient-level performance averaged across the five individual cross-validation models, whereas CV ensemble denotes performance after voxel-wise averaging of their predictions. $d_{\mathrm{SAM}}$ and LPIPS values are multiplied by 100 for readability. Complete region-wise results and statistical comparisons are provided in Appendix~\ref{app:perceptual_results}.}
\label{tab:sam_perceptual_compact}
\begin{tabular}{lllcccccc}
\toprule
& & & \multicolumn{3}{c}{MAE supervision}
& \multicolumn{3}{c}{SAM supervision} \\
\cmidrule(lr){4-6} \cmidrule(lr){7-9}
Task & Set & Prediction
& MAE $\downarrow$ & $d_{\mathrm{SAM}}\downarrow$ & LPIPS $\downarrow$
& MAE $\downarrow$ & $d_{\mathrm{SAM}}\downarrow$ & LPIPS $\downarrow$ \\
\midrule
\multirow{4}{*}{Task 1}
& \multirow{2}{*}{AB/HN/TH}
& Mean CV
& \textbf{70.75} & 24.27 & 8.34
& 78.26 & \textbf{18.99} & \textbf{7.06} \\
&
& CV ensemble
& \textbf{66.98} & 24.31 & 8.22
& 73.21 & \textbf{19.85} & \textbf{7.03} \\
\cmidrule(lr){2-9}
& \multirow{2}{*}{Ext-T2}
& Mean CV
& 104.39 & 37.43 & 13.82
& \textbf{96.36} & \textbf{33.71} & \textbf{12.00} \\
&
& CV ensemble
& 97.43 & 37.30 & 13.25
& \textbf{90.14} & \textbf{35.93} & \textbf{11.79} \\
\midrule
\multirow{4}{*}{Task 2}
& \multirow{2}{*}{AB/HN/TH}
& Mean CV
& \textbf{64.95} & 17.54 & 5.48
& 69.52 & \textbf{13.99} & \textbf{4.69} \\
&
& CV ensemble
& \textbf{61.28} & 17.88 & 5.40
& 65.18 & \textbf{14.36} & \textbf{4.59} \\
\cmidrule(lr){2-9}
& \multirow{2}{*}{Sim-CBCT}
& Mean CV
& 114.78 & 20.87 & 6.34
& \textbf{109.23} & \textbf{17.94} & \textbf{5.31} \\
&
& CV ensemble
& 111.88 & 20.67 & 6.15
& \textbf{106.33} & \textbf{17.99} & \textbf{5.15} \\
\bottomrule
\end{tabular}

\end{table*}
\renewcommand{\arraystretch}{1}

In the Ext-T2 and Sim-CBCT settings, SAM supervision improves MAE, $d_{\mathrm{SAM}}$, and LPIPS for both tasks and both prediction strategies. Ensemble averaging primarily reduces MAE, whereas its effect on perceptual metrics is smaller and not uniformly favorable. This is consistent with voxel-wise averaging reducing random intensity errors while potentially attenuating fine structural details.

\subsection{Anatomical consistency}

The objective of this section is to analyze the extent to which the different registration strategies preserve the anatomy of the input image in the synthesized outputs. 
To highlight these differences, we introduce a new anatomy-oriented similarity metric. The metric is derived from feature representations extracted by the pretrained TS CT 3\,mm segmentation model (M297) used within IMPACT-Reg. The feature extractor remains fixed during evaluation. The underlying intuition is that a similarity metric capable of accurately assessing multimodal anatomical correspondence in image registration should also provide a relevant measure of anatomical consistency between the input image and the synthesized image.

Anatomical consistency results reported in Table~\ref{tab:task2_cv_0_reg_reg_region_metrics} show substantial differences in the deformation required to align the synthesized images with the input CBCT, as measured with the IMPACT-Reg metric M297. Smaller deformations, i.e., higher anatomical consistency, are consistently obtained for IMPACT-trained models across all evaluated regions.

In in-distribution regions, the relative difference between ELX and IMPACT reaches 124.25\% in the abdominal region and 136.65\% in head-and-neck cases (both $p<0.001$), indicating substantially larger deformations, and thus lower anatomical consistency, for ELX-trained models. In the thoracic region, the difference remains statistically significant but is markedly smaller (3.59\%, $p<0.01$). Aggregated across AB/HN/TH regions, ELX-trained models exhibit an overall relative difference of 81.82\% compared to IMPACT-trained models ($p<0.001$).

A similar trend is observed on Sim-CBCT, where ELX-trained models still require significantly larger deformations than IMPACT-trained models, with a relative difference of 9.65\% ($p<0.001$).

Together, these results show that IMPACT-trained models better preserve the anatomy of the input CBCT in both real and simulated CBCT settings.

\begin{table}[h!]
\centering
\footnotesize
\setlength{\tabcolsep}{6pt}
\renewcommand{\arraystretch}{1.25}
\caption{Registration-based deformation measure between the input CBCT and the synthesized image for Task 2, using the IMPACT-Reg metric M297. Values report the relative percentage difference between ELX and IMPACT. Positive values indicate larger deformation for ELX compared to IMPACT. Results are reported across anatomical regions for models trained with the SAM criterion. Statistical significance is assessed using paired Wilcoxon signed-rank tests on matched patients.
}
\label{tab:task2_cv_0_reg_reg_region_metrics}
\begin{tabular}{lc}
\toprule
Region & IMPACT vs ELX \\
\midrule
AB  & 124.25\%$^{***}$ \\
HN & 136.65\%$^{***}$ \\
TH  & 3.59\%$^{**}$ \\
AB/HN/TH & 81.82\%$^{***}$ \\
\addlinespace[0.25em]
\hline
\hline
\addlinespace[0.25em]
Sim-CBCT  & 9.65\%$^{***}$ \\
\bottomrule
\end{tabular}

\end{table}
\renewcommand{\arraystretch}{1}

\subsection{Official SynthRAD results as a benchmark-bias case study}

The official SynthRAD evaluation provides an external case study of registration-dependent benchmark behavior. In the local experiments, IMPACT-aligned supervision is associated with higher measured anatomical consistency and better performance in several registration-independent or better-aligned settings. In contrast, the official server favors ELX-trained models, whose targets are more consistent with the registration convention used to construct the challenge evaluation references.

Table~\ref{tab:public_split} illustrates this reversal. Under the official protocol, ELX training improves all reported metrics relative to IMPACT training in both tasks. For example, MAE decreases from 75.82 to 68.20 HU in Task~1 and from 56.05 to 52.87 HU in Task~2. These results do not contradict the local anatomical analyses; rather, they show that reference-based scores reflect both synthesis quality and compatibility with the geometry embedded in the evaluation data.

\begin{table}[h!]
\centering
\footnotesize
\caption{ELX- versus IMPACT-trained models on the public SynthRAD2025 validation set, as scored by the challenge server. Higher ELX scores reflect agreement with the ELX-consistent evaluation geometry, not established source-anatomy preservation.}
\label{tab:public_split}
\begin{tabular}{lcc|cc}
\toprule
\multirow{2}{*}{\textbf{Metric}} & \multicolumn{2}{c|}{\textbf{Task 1}} & \multicolumn{2}{c}{\textbf{Task 2}} \\
 & \textbf{ELX} & \textbf{IMPACT} & \textbf{ELX} & \textbf{IMPACT} \\
\midrule
MAE  & 68.20 & 75.82 &       52.87 & 56.05 \\
PSNR & 29.81 & 28.70 &  32.36 & 31.65 \\
SSIM & 0.92 & 0.91 &        0.96 & 0.95 \\
Dice & 0.72 & 0.70 &        0.83 & 0.82 \\
HD95 & 8.42 & 8.89 &         5.40 & 5.41 \\
\bottomrule
\end{tabular}

\end{table}

The final challenge submission therefore used the five-fold ELX ensemble trained with SAM-based perceptual supervision. This
configuration was selected because the public validation results favored ELX-consistent training, while SAM supervision improved
downstream Dice and qualitative structural preservation despite a moderate increase in local voxel-wise MAE. Checkpoints within each configuration remained selected exclusively according to validation MAE. For the final submission only, the global model was further fine-tuned into two region-specific sub-models (AB+TH and HN); all other results in this paper use the global models.

The submitted method ranked third overall for both MR$\rightarrow$CT and CBCT$\rightarrow$CT synthesis. It achieved MAEs of 67.24 and 53.09 HU, Dice scores of 0.737 and 0.843, and HD95 values of 7.51 and 5.08 mm, respectively. Dosimetric performance was also competitive, including high gamma pass rates. Complete image-based and dosimetric rankings are provided in Appendix~\ref{app:challenge_rankings}.

Together, the public validation reversal and the official test results show that the proposed models are competitive under the prescribed challenge protocol while highlighting a limitation of reference-based ranking. The fact that ELX is favored by the ELX-consistent server, whereas IMPACT is favored by several source-preservation and OOD analyses, demonstrates that the evaluation convention can influence the apparent ordering of synthesis strategies.

\subsection{Estimation of registration-induced sensitivity in sCT evaluation metrics}

These experiments estimate the magnitude of evaluation error that can arise from residual multimodal misregistration alone, without any synthesis process. Rather than defining a theoretical performance bound, the resulting values provide an empirical registration-induced reference level for MAE, PSNR, SSIM, $d_{\mathrm{SAM}}$, LPIPS, and Dice.

To isolate this effect from any synthesis error, all controls were constructed from CT images only. The goal was to ask how much the evaluation metrics can change when the CT reference geometry is modified by registration, even though no synthetic CT is generated. The control was adapted to the evaluation geometry available in each benchmark. In SynthRAD2023 brain and pelvis cases, where the official CT reference is only rigidly aligned to the MR image, we compared this rigid CT with an IMPACT-deformed CT to estimate the effect of adding a plausible non-rigid correspondence. In SynthRAD2025 AB, HN, and TH cases, where deformable registration is already used, we applied a cycle-deformation control: the CT was first transformed according to the ELX deformation and then mapped back using the IMPACT deformation field. This estimates the metric sensitivity to switching between two plausible non-rigid registration conventions, without introducing any synthesis model.

This analysis uses the disagreement between ELX and IMPACT as a proxy for registration-dependent geometric uncertainty. The preceding segmentation analysis indicates higher average anatomical consistency for IMPACT, but neither method is treated as ground truth. Agreement between their deformation fields suggests a relatively well-constrained correspondence, whereas disagreement identifies regions in which the estimated anatomy and the resulting evaluation metrics are more sensitive to the registration convention.

Table~\ref{tab:registration-supervised-comparison-all} compares these CT-derived controls with representative supervised sCT results. In the SynthRAD2025 AB/HN/TH regions, registration alone produces aggregated MAEs of 60.02 HU for Task~1 and 53.42 HU for Task~2, with corresponding SSIM values of 0.927 and 0.938. In the SynthRAD2023 brain/pelvis regions, the estimated MAEs are 59.18 HU and 39.50 HU for Tasks~1 and 2, respectively. These values are of the same order of magnitude as those obtained by high-performing supervised synthesis methods, despite the absence of synthesis error in the CT-derived controls.

The regional variation is consistent with the dependence of intensity-based metrics on local image gradients. For a small residual
displacement $d(x)$, the induced intensity difference can be approximated by:
\begin{equation}
    \left|\nabla I(x)\cdot d(x)\right|.
\end{equation}
Consequently, comparable geometric errors can produce different MAEs across anatomical regions. Interfaces involving air, bone, teeth, or thin soft-tissue boundaries are particularly sensitive, helping to explain the larger metric variations observed in regions such as HN and pelvis.

For the aggregate comparisons, the supervised MAE, PSNR, and SSIM/MS-SSIM values correspond to the best official SynthRAD2023 and
SynthRAD2025 submissions. Regional image-similarity values for AB, HN, and TH are obtained from the best local validation results of the region-specific fine-tuned ELX/SAM models submitted to the challenge \cite{boussot2025registration}. The perceptual values are derived from the ELX-based experiments reported in Tables~\ref{tab:task1_elx_fold_vs_cv_mae_median_perceptual} and~\ref{tab:task2_elx_fold_vs_cv_mae_median_perceptual}, whereas Dice
uses the best result among the MAE-, VGG-, and SAM-trained models.

Several supervised results match or even outperform the CT-derived controls on MAE, PSNR, or SSIM, but this does not invalidate the sensitivity estimate. A synthesis model optimized against an imperfect reference can reduce voxel-wise penalties through local smoothing, attenuation of high-gradient boundaries, or adaptation to the evaluation geometry, whereas the CT-derived controls retain sharp CT structures. Consistent with this interpretation, the controls generally preserve substantially higher Dice scores, and their perceptual values are often closer to those of SAM-trained models than to purely MAE-trained models.

These estimates should be conservatively interpreted. The cycle experiment captures only the disagreement between two regularized and anatomically plausible registration methods, while the SynthRAD2023 deformations were intentionally constrained, particularly for brain cases. Additional perturbation experiments also showed that MAE increases rapidly with small changes to the deformation fields. Therefore, the controls likely underestimate the full effect of correspondence uncertainty.

Overall, high-performing supervised sCT methods operate within the same metric range as that induced by residual registration uncertainty alone. This suggests that current reference-based benchmarks may be approaching a registration-dependent performance ceiling, where small improvements partly reflect better adaptation to the evaluation geometry rather than improved patient-specific anatomical fidelity.

\definecolor{metricorange}{RGB}{230,126,34}
\newcommand{\betterfloor}[1]{\textcolor{red}{#1}}
\newcommand{\worsefive}[1]{\textcolor{metricorange}{#1}}

\begin{table*}[h!]
\centering
\scriptsize
\setlength{\tabcolsep}{4pt}
\renewcommand{\arraystretch}{1.12}
\caption{Registration-induced metric sensitivity (Reg., CT-derived controls) versus representative supervised sCT performance (Sup.). Red: supervised result better than the control. Orange: worse by more than 10\%. Uncolored: within 10\%.}
\label{tab:registration-supervised-comparison-all}
\begin{tabular}{lllcccccc}
\toprule
Task & Region & Type & MAE $\downarrow$ & PSNR $\uparrow$ & SSIM $\uparrow$ & $d_{\mathrm{SAM}}$ $\downarrow$ & LPIPS $\downarrow$ & Dice $\uparrow$ \\
\midrule

\multirow{10}{*}{Task 1}
& \multirow{2}{*}{AB} & Reg.
& $66.45 \pm 13.58$ & $28.34 \pm 1.38$ & $0.908 \pm 0.027$ & $28.448 \pm 6.007$ & $10.8 \pm 3.5$ & $0.839 \pm 0.040$ \\
& & Sup.
& \betterfloor{$64.89$} & \betterfloor{$29.10$} & \betterfloor{$0.91$} & \betterfloor{$26.70$} & \betterfloor{$10.7$} & $0.777$ \\

& \multirow{2}{*}{HN} & Reg.
& $62.96 \pm 13.87$ & $29.43 \pm 1.71$ & $0.953 \pm 0.020$ & $11.342 \pm 2.975$ & $2.5 \pm 0.8$ & $0.818 \pm 0.072$ \\
& & Sup.
& $65.15$ & \betterfloor{$30.20$} & $0.94$ & $11.96$ & \worsefive{$3.1$} & \worsefive{$0.731$} \\

& \multirow{2}{*}{TH} & Reg.
& $53.00 \pm 13.30$ & $30.71 \pm 2.13$ & $0.950 \pm 0.014$ & $21.936 \pm 5.210$ & $7.1 \pm 2.4$ & $0.803 \pm 0.046$ \\
& & Sup.
& \worsefive{$60.07$} & \betterfloor{$30.76$} & $0.94$ & \betterfloor{$20.50$} & $7.2$ & \worsefive{$0.706$} \\

& \multirow{2}{*}{AB/HN/TH} & Reg.
& $60.02 \pm 13.85$ & $29.28 \pm 1.85$ & $0.927 \pm 0.035$ & $20.736 \pm 8.527$ & $6.8 \pm 0.042$ & $0.820 \pm 0.056$ \\
& & Sup.
& $64.81 \pm 21.25$ & \betterfloor{$29.997 \pm 2.759$} & \betterfloor{$0.936 \pm 0.050$} & \betterfloor{$19.85 \pm 7.26$} & $7.0 \pm 3.8$ & $0.738 \pm 0.055$ \\

\cmidrule(lr){2-9}

& \multirow{2}{*}{Brain/Pelvis} & Reg.
& $59.18 \pm 12.50$ & $28.96 \pm 1.52$ & $0.914 \pm 0.038$ & $14.397 \pm 8.459$ & $4.8 \pm 4.1$ & $0.886 \pm 0.060$ \\
& & Sup.
& \betterfloor{$58.83 \pm 13.41$} & \betterfloor{$29.61 \pm 1.79$} & $0.885 \pm 0.029$ & -- & -- & -- \\

\midrule

\multirow{10}{*}{Task 2}
& \multirow{2}{*}{AB} & Reg.
& $55.88 \pm 13.86$ & $29.78 \pm 1.84$ & $0.930 \pm 0.025$ & $17.348 \pm 4.548$ & $5.5 \pm 2.1$ & $0.745 \pm 0.058$ \\
& & Sup.
& $58.46$ & \betterfloor{$31.33$} & $0.90$ & $18.10$ & \worsefive{$6.6$} & \worsefive{$0.670$} \\

& \multirow{2}{*}{HN} & Reg.
& $69.50 \pm 22.42$ & $28.46 \pm 2.50$ & $0.945 \pm 0.023$ & $10.198 \pm 1.967$ & $2.0 \pm 0.6$ & $0.749 \pm 0.082$ \\
& & Sup.
& \betterfloor{$60.97$} & \betterfloor{$30.38$} & $0.94$ & \betterfloor{$9.50$} & $2.1$ & $0.720$ \\

& \multirow{2}{*}{TH} & Reg.
& $52.87 \pm 16.63$ & $30.96 \pm 2.57$ & $0.932 \pm 0.019$ & $16.532 \pm 3.741$ & $4.8 \pm 1.7$ & $0.784 \pm 0.066$ \\
& & Sup.
& \betterfloor{$50.40$} & \betterfloor{$31.78$} & $0.92$ & \betterfloor{$16.12$} & \worsefive{$5.4$} & $0.720$ \\

& \multirow{2}{*}{AB/HN/TH} & Reg.
& $53.42 \pm 20.18$ & $30.46 \pm 2.82$ & $0.938 \pm 0.031$ & $14.510 \pm 4.793$ & $4.0 \pm 2.2$ & $0.759 \pm 0.072$ \\
& & Sup.
& \betterfloor{$48.28 \pm 13.35$} & \betterfloor{$32.619 \pm 2.307$} & \betterfloor{$0.968 \pm 0.025$} & \betterfloor{$14.36 \pm 5.10$} & \worsefive{$4.6 \pm 2.6$} & $0.700 \pm 0.082$ \\

\cmidrule(lr){2-9}

& \multirow{2}{*}{Brain/Pelvis} & Reg.
& $39.50 \pm 13.09$ & $32.15 \pm 2.61$ & $0.942 \pm 0.043$ & $12.625 \pm 9.736$ & $4.0 \pm 4.6$ & $0.866 \pm 0.089$ \\
& & Sup.
& \worsefive{$49.95 \pm 11.78$} & $30.79 \pm 2.00$ & $0.906 \pm 0.036$ & -- & -- & -- \\

\bottomrule
\end{tabular}

\end{table*}

\section{Discussion} \label{sec:discussion_bias}

The results reported above support a coherent interpretation centered on the interaction between registration quality, voxel-wise supervision, and the metrics used to assess synthesis performance. 
This view directly builds on the concept of registration-induced target uncertainty introduced in Section~\ref{sec:i2i_background}, where residual registration errors were described as structured geometric label noise shaping the empirical conditional distribution $p(\tilde{y}\mid x)$ learned by supervised synthesis models.

\subsection{Registration bias as structured label noise}

The central finding of this study is that reference-based synthesis performance depends not only on the synthesis model, but also on the compatibility between the registration conventions used to construct the training targets and evaluation references. Residual registration errors therefore act as structured geometric label noise at two levels: they shape the target distribution learned during training and influence the reference against which predictions are subsequently scored \cite{dohmen2024similarity}.

The matched and mismatched experiments provide direct evidence for this effect. As summarized in Table~\ref{tab:registration_consistency_aggregate}, evaluating IMPACT-trained models against ELX rather than IMPACT references increases aggregate MAE from 63.55 to 72.47 HU in Task~1 and from 58.76 to 69.17 HU in Task~2. ELX-trained models similarly perform better under the ELX convention in both tasks. Because the predictions remain unchanged while only the evaluation reference is replaced, these differences show that voxel-wise scores measure both synthesis fidelity and agreement with the selected registration geometry.

Registration inconsistency also affects what the model learns. The qualitative example in Figure~\ref{fig:sct_mr_comparison_synthesis} shows blurred and displaced interfaces around the diaphragm in the
ELX-trained prediction. As described in Section~\ref{sec:i2i_background}, spatially variable targets broaden
the observed distribution $p(\tilde{y}\mid x)$. Under voxel-wise $\ell_1$ optimization, predicting a smoother intermediate boundary can reduce the expected penalty associated with placing a sharp structure at an uncertain location. Thus, although architectures with spatial skip connections possess a strong inductive bias toward preserving spatial organization, the optimization objective ultimately dominates the learned behavior. Under inconsistent voxel-wise supervision, ERM drives the model toward solutions that minimize the expected reconstruction error, even when this requires smoothing anatomical boundaries, attenuating high-frequency structures, or locally altering patient-specific anatomy.

The CT-derived controls in Table~\ref{tab:registration-supervised-comparison-all} reinforce this interpretation. High-performing supervised models operate close to the metric variation induced by residual registration alone and sometimes obtain better MAE, PSNR, or SSIM values than the controls. This does not imply superior anatomical fidelity. Unlike a sharp deformed CT, a voxel-wise optimized model can adapt to the evaluation target through local smoothing, attenuation of uncertain boundaries, or reproduction of its geometric convention. The substantially higher Dice scores of the CT-derived controls support this distinction: better intensity
agreement with a registered reference does not necessarily correspond to better structural preservation.

This interpretation is consistent with the subsequent study by Zimmermann et al.~\cite{zimmermann2026eliminating}, which used physics-based CBCT simulation to generate geometrically aligned pairs and IMPACT registration for real CBCT-CT data \cite{boussot2025impact}. Their findings similarly indicate that reducing registration-related inconsistencies improves anatomical and geometric coherence, even when conventional intensity metrics do not show a corresponding improvement.

More generally, benchmark rankings should be interpreted as protocol-dependent measurements rather than absolute indicators of anatomical accuracy or clinical superiority. The reversal between the local source-preservation analyses and the ELX-consistent SynthRAD server shows that small improvements in reference-based metrics may partly reflect adaptation to the benchmark geometry. Reliable sCT evaluation should therefore document the registration convention and complement voxel-wise scores with anatomy-oriented and task-specific criteria.

\subsection{Anatomical consistency and source preservation} 

The registration-induced bias identified above is not limited to voxel-wise image similarity metrics. It should also affect the anatomical relationship between the source CBCT and the synthesized CT, since a model trained on imperfectly aligned targets may learn to reproduce the geometry of the registered CT reference rather than preserve the patient-specific anatomy of the input image.

This interpretation is supported by the anatomical consistency analysis reported in Table~\ref{tab:task2_cv_0_reg_reg_region_metrics}. The IMPACT-Reg metric estimates the amount of deformation required to bring the synthesized image into anatomical agreement with the source CBCT representation. Larger values therefore indicate lower anatomical consistency with the input image. Across all evaluated regions, ELX-trained models require larger deformations than IMPACT-trained models, indicating that the synthesized images produced after ELX-based supervision deviate more strongly from the source anatomy.

The effect is particularly pronounced in the abdominal and head-and-neck regions, where the relative differences between ELX- and IMPACT-trained models exceed 120\%. These regions are characterized by complex soft-tissue structures and larger residual multimodal registration uncertainty, making them especially sensitive to geometric bias in the supervised targets. In the thoracic region, the difference is smaller but remains statistically significant, suggesting that the magnitude of the effect depends on the anatomical region and the difficulty of establishing reliable multimodal correspondences.

This analysis provides complementary evidence that improving the anatomical consistency of the training pairs reduces the propagation of geometric bias into the synthesized images. This effect is measured with respect to the source CBCT rather than only against the registered CT reference. It therefore directly supports the central objective of anatomy-preserving synthesis: the generated CT-like image should adapt appearance toward the CT domain without altering the patient-specific anatomy present in the input image.

Together, these results show that agreement with a registered CT reference is insufficient to assess anatomical validity. Registration-based anatomical consistency metrics provide a useful complementary evaluation criterion, particularly for downstream tasks including segmentation and deformable registration, where preserving source anatomy is more important.

\subsection{Consequences on uncertainty and OOD generalization}

Residual registration errors may also affect predictive uncertainty. The variability measured across models and test-time augmentations is commonly associated with epistemic uncertainty, but, under spatially inconsistent supervision, it may additionally reflect sensitivity to the structured geometric noise present in the training targets. Different models can therefore converge toward slightly different solutions of the biased conditional distribution $p(\tilde{y}\mid x)$, even when the underlying synthesis mapping is otherwise well constrained. 

From this perspective, the measured uncertainty is not purely epistemic. It is partly driven by structured geometric label noise introduced by residual registration errors. Because independently trained models are exposed to different empirical realizations of this spatial inconsistency, they may converge toward slightly different conditional solutions, thereby increasing inter-model variability. 

Table~\ref{tab:cv_uncertainty_mae_percentage} supports this interpretation: ELX-trained models exhibit higher predictive variability than IMPACT-trained models in almost all regions. In Task~1, the increase reaches 41.2\% in the abdomen and 29.7\% across the in-distribution AB/HN/TH regions; it rises to 116.1\% in the OOD pelvis and 61.8\% on Ext-T2. Task~2 shows the same, although weaker, trend, with increases of 27.6\% in the abdomen and 12.5\% across the in-distribution regions. Thus, the registration convention with higher measured anatomical consistency is also associated with more stable predictions.

The OOD results further indicate that this effect extends beyond agreement with a particular evaluation geometry. The OOD references are independent of the registration conventions used for training and rely on either expert-refined or registration-free correspondences. Nevertheless, IMPACT-trained models consistently outperform ELX-trained models. On Ext-T2, MAE decreases from 97.43 to 88.54 HU ($p<0.001$). On Sim-CBCT, it decreases from 71.09 to 60.41 HU in the abdomen, from 149.66 to 136.18 HU in the head and neck, and from 109.15 to 95.99 HU in the thorax, with corresponding improvements in PSNR and SSIM (Table~\ref{tab:ood_combined}).

Together, these findings suggest that anatomically more consistent supervision reduces sensitivity to dataset-specific geometric variations and acts as a form of regularization under distribution shift. Conversely,
noisier correspondences may encourage adaptation to the spatial conventions of the training set, increasing predictive variability and reducing OOD robustness.

\subsection{Origin of blurring under voxel-wise ERM}

The results support the interpretation introduced in Section~\ref{sec:i2i_background}: voxel-wise blurring in supervised synthesis is not only caused by intrinsic modality ambiguity, but also by residual geometric uncertainty in the registered targets. When training references contain spatially shifted structures, voxel-wise ERM favors intermediate anatomical configurations that reduce average reconstruction error without necessarily preserving the patient-specific source anatomy.

Perceptual, adversarial, or diffusion-based objectives can reduce this averaging effect by encouraging outputs that lie closer to the distribution of realistic CT images. However, these strategies do not remove the systematic bias of the supervised targets themselves: when trained on imperfectly aligned pairs, they still learn from $p(\tilde{y}\mid x)$ rather than from the ideal anatomical distribution $p(y\mid x)$. Their main benefit is therefore to improve sharpness and structural coherence, not to fully correct registration-induced bias.

This creates a direct tension with reference-based evaluation. When the reference geometry is imperfectly registered, MAE may favor smoothed predictions or registration-convention matching, even when sharper anatomy would be more desirable for downstream tasks such as segmentation or deformable registration.

\subsection{Why does SAM improve anatomical preservation?}

The consistent improvement obtained with SAM-based perceptual supervision over both MAE- and VGG-based objectives suggests that the choice of feature representation is critical for anatomy-preserving synthesis. Rather than comparing images voxel by voxel, perceptual supervision evaluates similarity in a learned feature space, where local structures are represented together with their surrounding anatomical context. As a result, the model is less encouraged to converge toward locally averaged intensity patterns and more encouraged to preserve coherent anatomical configurations.

When formulated as an $\ell_1$ loss in this feature space, the objective still estimates a conditional median, but over the representation induced by the encoder rather than over individual voxels. We hypothesize that this is the main driver of the observed improvement. The Hiera encoder compresses the image into low-resolution feature maps, so that each feature vector encodes a local anatomical configuration within its global context. The median would then be taken over spatially coherent configurations, a regime in which the optimal solution cannot be approximated by a blurred intensity average, which would implicitly constrain the generator to preserve sharp anatomical structures.

This also explains the advantage of SAM over VGG. Classification does not require preserving boundaries throughout the feature hierarchy, so VGG progressively discards spatial organization in favor of global appearance statistics. Segmentation does, and the SAM encoder retains boundary localization despite strong spatial compression.

The qualitative and downstream results support this interpretation. Models trained with voxel-wise losses produce smoother images that can remain
favorable under MAE, but often lose thin structures and sharp anatomical interfaces, particularly around pulmonary boundaries and the diaphragm (Fig.~\ref{fig:sam_comparison}). Perceptual supervision, especially with SAM features, produces sharper and more structurally coherent synthetic CT images, which in turn improves downstream segmentation performance. This illustrates a key limitation of voxel-wise evaluation: better agreement with an imperfect reference image does not necessarily imply better preservation of
patient-specific anatomy.

The comparison with CT-derived controls further supports this point. These controls isolate the discrepancy caused by residual deformation without
introducing synthesis error. The fact that SAM-trained predictions are closer to these controls in perceptual feature spaces suggests that SAM supervision moves synthetic CT images toward more realistic CT-like structural representations, whereas MAE optimization can remain competitive in
voxel-wise error while producing anatomically smoother images.

Importantly, the relationship between perceptual supervision and voxel-wise metrics changes when spatial correspondence becomes more reliable. In the registration-free Sim-CBCT setting, SAM supervision improves both perceptual and voxel-wise metrics: the ensemble MAE decreases from 111.88 to 106.33 HU, $d_{\mathrm{SAM}}$ from 20.67 to 17.99, and LPIPS from 6.15 to 5.15 (Table~\ref{tab:sam_perceptual_compact}). Under imperfect registration, voxel-wise $\ell_1$ objectives favor blurred spatial averages; under accurate correspondence, sharp and anatomically coherent structures are also spatially consistent with the reference, and the two criteria no longer conflict.

Overall, these findings suggest that anatomy- or realism-oriented synthesis methods should not be judged solely by their ability to outperform regression baselines on MAE when references are imperfectly registered. Their main contribution may lie in improving anatomical sharpness and structural coherence, with voxel-wise gains becoming visible mainly when the evaluation reference is geometrically reliable.

\subsection{Clinical relevance under task-specific constraints}

The impact of registration-induced bias should be interpreted in light of the intended clinical application. For radiotherapy dose calculation, the dependence on imperfect paired references may remain acceptable, provided that the generated attenuation map is sufficiently accurate. This is consistent with the official SynthRAD results (Tables~\ref{tab:cbctct_dose} and~\ref{tab:mrct_dose}), where BreizhCT achieved strong dose-based performance despite the limitations identified in the image-based analysis.

However, this conclusion does not directly extend to all downstream tasks. Dose metrics can be relatively tolerant to local blurring, small residual deformations, or the partial suppression of fine structures, because dose calculation depends primarily on attenuation properties. In contrast, broader domain adaptation tasks such as segmentation or deformable registration require stronger preservation of patient-specific anatomy, especially near thin structures and high-gradient interfaces.

These task-dependent requirements suggest that increasingly complex generative models should be motivated by a clearly identified clinical need,
rather than by visual sharpness alone. For dose calculation, voxel-wise accuracy may be sufficient in many cases; for anatomy-driven tasks, structural fidelity becomes essential.

\section{Conclusion}

This work shows that supervised synthetic CT generation is not only an image-to-image translation problem, but also a problem of supervision quality. When paired MRI--CT or CBCT--CT data are constructed through imperfect registration, the reference CT does not represent a true voxel-wise ground truth. Instead, residual misalignments introduce structured geometric label noise that can shape both model training and model evaluation.

Our results demonstrate that supervised models are sensitive to the registration convention used to define the training targets. Quantitative
performance improves when the same registration strategy is used for both training and evaluation, indicating that voxel-wise metrics can partly reward agreement with the registration pipeline rather than faithful preservation of the source anatomy. This effect also impacts OOD behavior and predictive uncertainty: models trained from less geometrically consistent targets show reduced robustness and larger prediction variability.

We further show that these effects are reduced when training uses the registration convention with higher measured anatomical consistency. IMPACT-aligned supervision is associated with better source-anatomy preservation, lower prediction variability, and improved OOD performance in the evaluated settings. However, IMPACT is not a ground-truth alignment, and both registration conventions may retain residual errors. More generally, changing the registration method does not remove the fundamental limitation of voxel-wise supervision: when the reference is imperfectly aligned, optimizing MAE, PSNR, or SSIM can favor smoothed or geometrically biased predictions.

To address part of this limitation, we introduced SAM-based perceptual supervision. Compared with MAE-only and VGG-based perceptual objectives,
SAM-based supervision improves downstream anatomical metrics and produces sharper, more structurally coherent synthetic CT images. This suggests that feature representations learned for segmentation provide a more appropriate supervisory signal for anatomy-preserving medical image synthesis than purely voxel-wise intensity losses.

Overall, our results argue that supervised synthetic CT generation should not be evaluated solely as an intensity regression problem. In the presence of imperfectly aligned references, voxel-wise metrics may reward smoothing or adaptation to the evaluation registration convention rather than true anatomical fidelity. This helps explain why highly optimized regression-based baselines remain difficult to outperform in challenge settings, and why sharper perceptual, generative, or unsupervised methods may appear quantitatively inferior despite producing more anatomically coherent images. Future benchmarks should therefore move beyond reference-based intensity metrics alone and include anatomy-oriented criteria that assess whether the synthesized CT preserves the patient-specific structures present in the input image.

\section{Data and Code Availability}
The IMPACT registrations used in this study are publicly released as Elastix B-spline transformation files, under a CC BY-NC 4.0 license. They can be applied with Transformix to the original SynthRAD images, which are not redistributed. Cases from centers whose data are restricted to challenge use are excluded.
\begin{itemize}
\item SynthRAD2023:\newline \url{https://huggingface.co/datasets/VBoussot/synthrad2023-impact-registration}
\item SynthRAD2025:\newline \url{https://huggingface.co/datasets/VBoussot/synthrad2025-impact-registration}
\end{itemize}
The code and configuration files needed to reproduce the submitted SynthRAD2025 solutions, implemented with KonfAI, are available for both tasks.
\begin{itemize}
\item Task~1:\newline \url{https://github.com/vboussot/Synthrad2025_Task_1}
\item Task~2:\newline \url{https://github.com/vboussot/Synthrad2025_Task_2}
\end{itemize}

\section*{Acknowledgment}
The work presented in this article was supported by the French National Research Agency as part of the VATSop project (ANR-20-CE19-0015). Additionally, it was funded by the French National Research Agency as part of the DIMADOSE project (C. Hémon). While preparing this work, the authors used ChatGPT to enhance the writing structure and refine grammar. After using these tools, the authors reviewed and edited the manuscript and take full responsibility for its content. The authors have no relevant financial or non-financial interests to disclose. 

\bibliographystyle{elsarticle-num} 
\bibliography{Biblio.bib}

\appendices

The appendices provide supplementary information supporting the main experiments. Appendix~\ref{app:synthrad} describes the SynthRAD
datasets, evaluation protocol, and official challenge rankings. Appendix~\ref{app:implementation} reports the complete implementation
details of the synthesis pipeline. Appendix~\ref{app:regional_results} contains the full region-wise results underlying the aggregate analyses presented in the main text.

\section{SynthRAD challenge and benchmark details}
\label{app:synthrad}

This appendix provides additional information about the SynthRAD2023 and SynthRAD2025 benchmarks used in this study. It summarizes the
challenge design, dataset composition, and evaluation procedures that are relevant to the interpretation of registration-dependent
performance. The official image-based and dosimetric rankings of the submitted method are also reported for completeness.

The MICCAI Challenge SynthRAD combines multi-center datasets with a standardized evaluation framework for comparing synthetic CT generation from MRI and CBCT data \cite{thummerer2025synthrad2025,huijben2024generating}. Its two tasks address MR-to-CT synthesis for MR-guided radiotherapy and CBCT-to-CT synthesis for adaptive radiotherapy.

\subsection{Challenge design and objectives}

The challenge is organized as an open benchmarking platform, where participants are required to submit their methods in the form of containerized algorithms (Docker), ensuring full reproducibility and preventing any manual intervention during evaluation. 
This design enforces a strict separation between training and testing data, and enables standardized and fair comparison across competing approaches~\cite{thummerer2025synthrad2025}.

Two tasks are defined, reflecting distinct clinical scenarios:
\begin{itemize}
    \item \textbf{Task 1: MRI-to-CT synthesis}, targeting MR-only and MR-guided radiotherapy workflows.
    \item \textbf{Task 2: CBCT-to-CT synthesis}, targeting CBCT-based adaptive radiotherapy.
\end{itemize}

Participants may submit models for one or both tasks, and are required to handle all anatomical regions within a given task using either a unified or region-specific strategy.

The ground-truth CT images of the validation and test sets are not directly accessible to participants. 
Instead, submitted algorithms are executed on hidden data, and predictions are evaluated remotely through a centralized evaluation server. 
The challenge is further structured into multiple phases (training, validation, and test), each associated with strict submission limits, thereby reducing the risk of iterative overfitting to the evaluation data.

At the same time, the organizers provide a highly transparent description of the evaluation pipeline, including preprocessing, registration procedures, and quantitative metrics~\cite{thummerer2025synthrad2025}. While this improves reproducibility, it also facilitates metric-oriented optimization, where methods are progressively adapted to the benchmark criteria.

\subsection{Dataset characteristics}

The experiments combine SynthRAD2025 data for the primary in-distribution analyses with SynthRAD2023 data for complementary brain and pelvis evaluation. The two datasets differ in cohort composition, anatomical coverage, and evaluation registration, thereby providing complementary settings for studying registration-dependent behavior.

The challenge relies on the SynthRAD2025 dataset, which comprises 2362 patient cases collected from five European university medical centers, including 890 MRI-CT pairs and 1472 CBCT-CT pairs. 
The benchmark addresses two synthesis tasks: MRI-to-CT synthesis (Task~1) and CBCT-to-CT synthesis (Task~2), across three anatomical regions corresponding to common radiotherapy indications: head-and-neck (HN), thorax (TH), and abdomen (AB).

The dataset is divided into training (65\%), validation (10\%), and test (25\%) subsets. 
Only the training data are fully accessible to participants, while the validation and test sets remain partially hidden.

In addition to the official SynthRAD2025 benchmark, we conduct complementary local experiments using the SynthRAD2023 dataset~\cite{huijben2024generating} as an OOD evaluation set. 
This dataset contains 1080 paired MRI-CT and CBCT-CT acquisitions collected from three Dutch university medical centers, and covers two anatomical regions: brain and pelvis.

Overall, both datasets exhibit substantial multi-center and multi-protocol variability. 
Images are acquired using different scanners, acquisition settings, and clinical workflows, introducing significant inter-domain heterogeneity across patients and institutions. 
Although this variability increases the difficulty of the synthesis task, it also promotes the development of more robust and clinically generalizable models.

\subsection{Evaluation protocol}

The official evaluation combines image similarity, segmentation-based anatomical agreement, and dosimetric accuracy. Image similarity is assessed using MAE, PSNR, and MS-SSIM between sCT and CT. Geometric agreement is evaluated from TotalSegmentator structures using Dice and Hausdorff distance, after resampling to 3\,mm resolution \cite{wasserthal2023totalsegmentator}. Clinical relevance is assessed using dose error, dose-volume histogram differences, and gamma pass rates.

A key distinction concerns the registration applied before metric computation. Participants receive rigidly aligned multimodal pairs in both challenge editions. SynthRAD2025 additionally applies deformable multimodal registration during evaluation, whereas SynthRAD2023 relies on rigid alignment only. This difference motivates the separate analyses of the two benchmark settings in the main text.

\begin{figure*}[h!]
    \centering
  \includegraphics[width=0.8\textwidth]{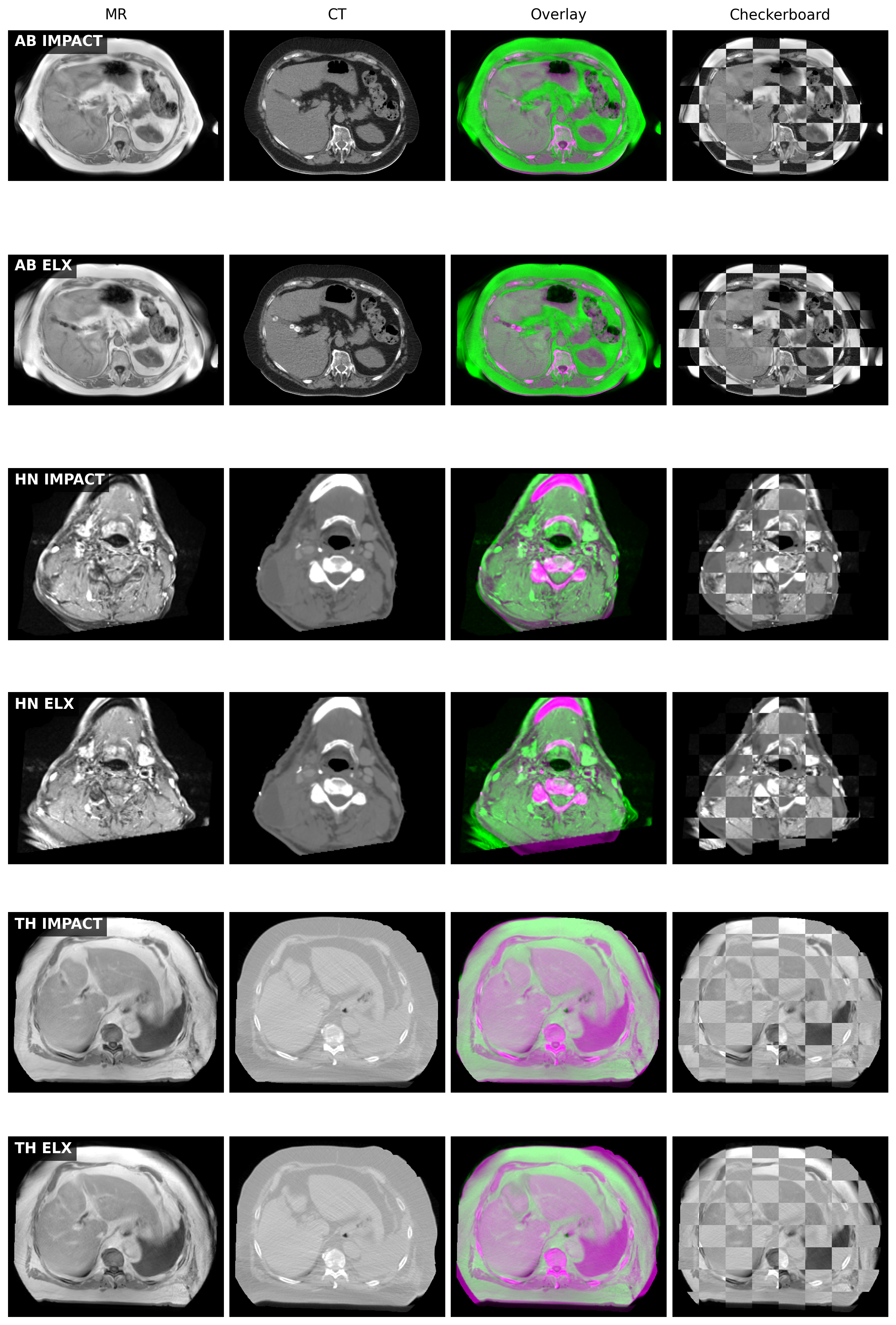}
\caption{Qualitative comparison of registration results between IMPACT and ELX across anatomical regions (Abdomen (AB), Head-and-Neck (HN), Thorax (TH)). For each case, the fixed image (MR), moving image (CT), overlay visualization, and checkerboard fusion are shown. In the displayed cases, IMPACT-based registration shows sharper and more spatially consistent correspondences in several boundary regions, particularly at soft-tissue interfaces, whereas ELX shows more visible local discrepancies. Checkerboard views further highlight structural inconsistencies.}
\label{fig:sct_mr_comparison}
\end{figure*}

\subsection{Official challenge rankings}
\label{app:challenge_rankings}

Tables~\ref{tab:mrct_image} and~\ref{tab:cbctct_image} report the official image-based rankings for the MR-to-CT and CBCT-to-CT tasks, respectively. The corresponding dosimetric rankings are provided in Tables~\ref{tab:mrct_dose} and~\ref{tab:cbctct_dose}. These results document the competitiveness of the submitted method under the prescribed challenge protocol; their relation to registration-dependent benchmark behavior is discussed in the main text.

\begin{table*}[h!]
\centering
\scriptsize
\renewcommand{\arraystretch}{1}
\setlength{\tabcolsep}{5pt}
\caption{
Official ranking for the MR$\rightarrow$CT synthesis task based on the challenge evaluation metrics. Performance is reported using commonly used image similarity metrics (MAE, PSNR, MS-SSIM) and segmentation-based metrics (Dice, HD95). The top 5 performing teams and the challenge baseline are shown. Our method (BreizhCT) ranks 3rd overall.
}
\label{tab:mrct_image}
\begin{tabular}{clccccc}
\toprule
\textbf{\#} & \textbf{Team} & \textbf{MAE$\downarrow$} & \textbf{PSNR$\uparrow$} & \textbf{MS-SSIM$\uparrow$} & \textbf{Dice$\uparrow$} & \textbf{HD95$\downarrow$} \\
\midrule
\textbf{1} & FelixSun (KoalAI)      & \textbf{64.81}  & \textbf{29.997} & \textbf{0.936} & 0.779 & 6.01 \\
\textbf{2} & JavierSequeiro         & 65.51  & 29.611 & 0.933 & 0.766 & 6.32 \\
\textbf{3} & \textcolor{cyan}{Valentin (BreizhCT)} & \textcolor{cyan}{67.24}  & \textcolor{cyan}{29.957} & \textcolor{cyan}{0.935} & \textcolor{cyan}{0.737} & \textcolor{cyan}{7.51} \\
\textbf{4} & siyuanmei (MixCT)      & 67.90  & 29.628 & 0.931 & 0.785 & 5.77 \\
\textbf{5} & hanbingocean (QWER)    & 75.68  & 28.756 & 0.922 & 0.715 & 7.69 \\
\midrule
\textbf{14} & MaartenTerpstra (baseline) & 309.28 & 18.630 & 0.466 & 0.006 & 136.34 \\
\bottomrule
\end{tabular}

\end{table*}

\begin{table*}[h!]
\centering
\scriptsize
\renewcommand{\arraystretch}{1}
\setlength{\tabcolsep}{4pt}
\caption{
Official ranking for the MR$\rightarrow$CT synthesis task based on dosimetric evaluation metrics. Performance is reported using dose-based metrics (Dose MAE, DVH) and gamma pass rate (GPR), for both $\gamma$ and $p$ criteria. The top 5 performing teams and the challenge baseline are shown. Our method (BreizhCT) ranks 3rd overall.
}
\label{tab:mrct_dose}
\begin{tabular}{clcccccc}
\toprule
\textbf{\#} & \textbf{Team} 
& \textbf{Dose MAE$_\gamma\downarrow$} 
& \textbf{Dose MAE$_p\downarrow$} 
& \textbf{DVH$_\gamma\downarrow$} 
& \textbf{DVH$_p\downarrow$} 
& \textbf{GPR$_\gamma\uparrow$} 
& \textbf{GPR$_p\uparrow$} \\
\midrule
\textbf{1} & FelixSun (KoalAI) 
& \textbf{0.006} & \textbf{0.024} & \textbf{0.011} & 0.064 & 98.33 & 84.04 \\
\textbf{2} & JavierSequeiro 
& 0.006 & 0.024 & 0.011 & \textbf{0.060} & 98.50 & 84.56 \\
\textbf{3} & \textcolor{cyan}{Valentin (BreizhCT)} 
& \textcolor{cyan}{0.006} & \textcolor{cyan}{0.027} & \textcolor{cyan}{0.013} & \textcolor{cyan}{0.067} & \textcolor{cyan}{98.88} & \textcolor{cyan}{82.19} \\
\textbf{4} & siyuanmei (MixCT) 
& 0.007 & \textbf{0.023} & 0.016 & 0.075 & 98.22 & 81.92 \\
\textbf{5} & hanbingocean (QWER) 
& 0.007 & 0.027 & 0.014 & 0.073 & 98.29 & 81.91 \\
\midrule
\textbf{14} & MaartenTerpstra (baseline) 
& 0.056 & 0.152 & 0.094 & 0.451 & 78.08 & 59.59 \\
\bottomrule
\end{tabular}

\end{table*}

For MR-to-CT synthesis, BreizhCT ranked third in both the image-based and dosimetric evaluations. The method remained close to the highest-ranked submissions across intensity, segmentation, and dose metrics, supporting its use as a competitive model in the analyses reported in the main text.

\begin{table*}[h!]
\centering
\scriptsize
\renewcommand{\arraystretch}{1}
\setlength{\tabcolsep}{5pt}
\caption{
Official ranking for the CBCT$\rightarrow$CT synthesis task based on challenge evaluation metrics. Performance is reported using commonly used image similarity metrics (MAE, PSNR, MS-SSIM) and segmentation-based metrics (Dice, HD95). The top 5 performing teams and the challenge baseline are shown. Our method (BreizhCT) ranks 3rd overall.
}
\label{tab:cbctct_image}
\begin{tabular}{clccccc}
\toprule
\textbf{\#} & \textbf{Team} & \textbf{MAE$\downarrow$} & \textbf{PSNR$\uparrow$} & \textbf{MS-SSIM$\uparrow$} & \textbf{Dice$\uparrow$} & \textbf{HD95$\downarrow$} \\
\midrule
\textbf{1} & GlassCity (MixCT)         & \textbf{48.27}  & \textbf{32.62} & \textbf{0.968} & \textbf{0.857} & \textbf{4.53} \\
\textbf{2} & JavierSequeiro            & 52.49  & 31.91 & 0.964 & 0.846 & 4.87 \\
\textbf{3} & \textcolor{cyan}{Valentin (BreizhCT)} 
& \textcolor{cyan}{53.09}  & \textcolor{cyan}{32.49} & \textcolor{cyan}{0.966} & \textcolor{cyan}{0.843} & \textcolor{cyan}{5.08} \\
\textbf{4} & ayuan (et)                & 53.61  & 31.89 & 0.963 & 0.842 & 4.99 \\
\textbf{5} & RicardoBrioso             & 62.75  & 31.01 & 0.952 & 0.801 & 6.65 \\
\midrule
\textbf{14} & MaartenTerpstra (baseline) & 308.99 & 18.77 & 0.505 & 0.004 & 133.48 \\
\bottomrule
\end{tabular}

\end{table*}

\begin{table*}[h!]
\centering
\scriptsize
\renewcommand{\arraystretch}{1}
\setlength{\tabcolsep}{4pt}
\caption{Official ranking for the CBCT$\rightarrow$CT synthesis task based on dosimetric evaluation metrics. Performance is reported using dose-based metrics (Dose MAE, DVH) and gamma pass rate (GPR), for both $\gamma$ and $p$ criteria. The top 5 performing teams and the challenge baseline are shown. Our method (BreizhCT) ranks 3rd overall.}
\label{tab:cbctct_dose}
\begin{tabular}{clcccccc}
\toprule
\textbf{\#} & \textbf{Team} 
& \textbf{Dose MAE$_\gamma\downarrow$} 
& \textbf{Dose MAE$_p\downarrow$} 
& \textbf{DVH$_\gamma\downarrow$} 
& \textbf{DVH$_p\downarrow$} 
& \textbf{GPR$_\gamma\uparrow$} 
& \textbf{GPR$_p\uparrow$} \\
\midrule
\textbf{1} & GlassCity (MixCT) 
& \textbf{0.004} & \textbf{0.017} & \textbf{0.013} & \textbf{0.034} & 99.300 & 88.640 \\
\textbf{2} & JavierSequeiro 
& 0.005 & 0.018 & 0.015 & 0.036 & 99.312 & 87.765 \\
\textbf{3} & \textcolor{cyan}{Valentin (BreizhCT)} 
& \textcolor{cyan}{0.005} & \textcolor{cyan}{0.020} & \textcolor{cyan}{0.015} & \textcolor{cyan}{0.036} & \textcolor{cyan}{99.308} & \textcolor{cyan}{86.407} \\
\textbf{4} & ayuan (et) 
& 0.005 & 0.018 & 0.015 & 0.039 & 99.227 & 87.405 \\
\textbf{5} & RicardoBrioso 
& 0.006 & 0.021 & 0.017 & 0.046 & 98.948 & 84.989 \\
\midrule
\textbf{14} & MaartenTerpstra (baseline) 
& 0.036 & 0.094 & 0.184 & 0.482 & 76.393 & 59.136 \\
\bottomrule
\end{tabular}

\end{table*}

For CBCT-to-CT synthesis, BreizhCT also ranked third overall. The submitted method achieved competitive image-similarity and anatomical metrics together with high gamma pass rates, confirming that the observed registration effects are not limited to a deliberately weak synthesis baseline.

\section{Implementation details}
\label{app:implementation}

This appendix reports the implementation details omitted from the main text for concision. The same preprocessing, architecture, optimization,
checkpoint-selection, and inference procedures were used across the registration and supervision configurations unless otherwise stated. Consequently, differences between configurations primarily reflect the registration convention and training objective (loss) rather than changes to the synthesis pipeline.

The supervised synthesis framework follows the pipeline illustrated in Fig.~\ref{fig:architecture}. It combines registration-based pairing, patch-based preprocessing, a 2.5D convolutional generator, and ensemble-based inference. All training, validation, and inference experiments were implemented within the KonfAI framework, which was used to configure the data pipeline, model architecture, optimization strategy, checkpoint selection, test-time augmentation, and ensemble inference in a reproducible manner \cite{boussot2025konfai}.

The synthesis model is based on a 2.5D U-Net++ architecture with a ResNet-34 encoder \cite{zhou2018unet++}. For each target axial slice, five adjacent slices are concatenated along the channel dimension, providing local through-plane context while preserving the computational efficiency of a 2D convolutional model. The decoder aggregates multi-scale features through the dense skip connections of U-Net++, and a final hyperbolic tangent activation constrains the output to the normalized CT intensity range. The generator contains 26,084,881 trainable parameters.

Training is performed on fixed-size in-plane patches of $320 \times 320$, using mini-batches of size 32 and random flipping augmentation. Models are optimized with AdamW using an initial learning rate of $10^{-3}$, momentum parameters $\beta_1 = 0.9$ and $\beta_2 = 0.999$, and a weight decay of $10^{-3}$. A step scheduler decreases the learning rate by a factor of $0.75$ every 10 epochs, with one epoch corresponding to 2500 training iterations. The final checkpoint is selected according to the lowest validation MAE and is typically obtained around 40,000 training iterations. This selection criterion is kept identical across all configurations so that the reported differences primarily reflect the effect of the registration and supervision strategies rather than differences in model selection.

At inference time, predictions are performed slice-wise using the same 2.5D implementation. Test-time augmentation is applied using flipping transformations, and predictions are averaged across augmentations. The outputs are then denormalized to recover CT intensities in Hounsfield units. For each evaluated configuration, the five models obtained from the cross-validation folds are ensembled at inference time, and final synthetic CT volumes are generated by averaging predictions across folds and augmentations.

\section{Complete region-wise results}
\label{app:regional_results}

This appendix provides the complete region-wise results underlying the aggregate comparisons reported in the main text. It includes patient-level means and standard deviations, statistical comparisons, and separate analyses for the registration conventions and training objectives.

\subsection{Registration consistency}
\label{app:registration_results}

Tables~\ref{tab:task1_cv_region_metrics_in_distribution} and~\ref{tab:task2_cv_region_metrics_in_distribution} provide the complete results for all combinations of training and evaluation registration conventions. The first column group indicates the registration used to construct the training targets, and the nested column groups indicate the registration used to define the evaluation reference. Brain and pelvis are included only for Task~1 in this comparison because they correspond to the complementary SynthRAD2023 setting.

\begin{table*}[h!]
\centering
\scriptsize
\setlength{\tabcolsep}{2pt}
\renewcommand{\arraystretch}{1.25}
\caption{Quantitative results for Task 1 (supervised cross-validation) across anatomical regions. Mean $\pm$ standard deviation of commonly used image similarity metrics (MAE, PSNR, SSIM) are reported. Column groups indicate the registration method used during training (IMPACT or ELX), while subgroups correspond to the registration method used for evaluation. Regions AB, HN, and TH correspond to in-distribution data, whereas brain and pelvis represent out-of-distribution regions not seen during training. Statistical comparisons against IMPACT/IMPACT configuration follow the patient-level protocol described in Section~\ref{sec:statistics}.}
\label{tab:task1_cv_region_metrics_in_distribution}
\begin{tabular}{lcccccccccccc}
\toprule
 & \multicolumn{6}{c}{IMPACT} & \multicolumn{6}{c}{ELX} \\
\cmidrule(lr){2-7} \cmidrule(lr){8-13}
Region & \multicolumn{3}{c}{IMPACT} & \multicolumn{3}{c}{ELX} & \multicolumn{3}{c}{IMPACT} & \multicolumn{3}{c}{ELX} \\
\cmidrule(lr){2-4} \cmidrule(lr){5-7} \cmidrule(lr){8-10} \cmidrule(lr){11-13}
 & MAE & PSNR & SSIM & MAE & PSNR & SSIM & MAE & PSNR & SSIM & MAE & PSNR & SSIM \\
\midrule
AB & \bmstd{58.89}{9.16}{} & \bmstd{29.94}{1.43}{} & \bmstd{0.909}{0.022}{} & \mstd{72.14}{13.78}{***} & \mstd{27.68}{1.57}{***} & \mstd{0.899}{0.032}{**} & \mstd{64.04}{8.78}{***} & \mstd{29.29}{1.23}{***} & \mstd{0.900}{0.025}{***} & \mstd{67.73}{14.25}{**} & \mstd{28.47}{1.73}{***} & \mstd{0.904}{0.032}{} \\
HN & \bmstd{76.21}{11.31}{} & \bmstd{28.56}{1.29}{} & \bmstd{0.931}{0.027}{} & \mstd{88.51}{28.29}{*} & \mstd{27.43}{2.74}{*} & \mstd{0.915}{0.043}{} & \mstd{82.10}{12.83}{***} & \mstd{28.00}{1.35}{***} & \mstd{0.924}{0.030}{***} & \mstd{78.98}{23.41}{} & \mstd{28.38}{2.41}{} & \mstd{0.925}{0.038}{} \\
TH & \mstd{56.44}{10.81}{} & \bmstd{31.29}{1.86}{} & \mstd{0.941}{0.018}{} & \mstd{58.14}{11.12}{} & \mstd{30.23}{1.74}{} & \mstd{0.940}{0.027}{} & \mstd{59.81}{9.85}{***} & \mstd{30.70}{1.56}{***} & \mstd{0.936}{0.022}{***} & \bmstd{55.30}{10.14}{} & \mstd{30.74}{1.66}{} & \bmstd{0.943}{0.028}{} \\
AB/HN/TH & \bmstd{63.55}{13.61}{} & \bmstd{29.97}{1.91}{} & \bmstd{0.927}{0.026}{} & \mstd{72.47}{22.68}{***} & \mstd{28.49}{2.42}{***} & \mstd{0.918}{0.038}{**} & \mstd{68.31}{14.27}{***} & \mstd{29.37}{1.78}{***} & \mstd{0.920}{0.030}{***} & \mstd{66.98}{19.27}{} & \mstd{29.24}{2.24}{**} & \mstd{0.924}{0.036}{} \\
\addlinespace[0.25em]
\hline
\hline
\addlinespace[0.25em]
Brain & \bmstd{114.55}{11.10}{} & \bmstd{24.81}{0.80}{} & \bmstd{0.858}{0.021}{} & \mstd{138.52}{16.24}{***} & \mstd{23.37}{0.98}{***} & \mstd{0.837}{0.023}{***} & \mstd{117.32}{10.47}{*} & \mstd{24.66}{0.73}{*} & \mstd{0.856}{0.020}{} & \mstd{140.32}{18.43}{***} & \mstd{23.35}{1.06}{***} & \mstd{0.836}{0.021}{***} \\
Pelvis & \bmstd{65.40}{13.08}{} & \bmstd{28.67}{1.75}{} & \bmstd{0.850}{0.051}{} & \mstd{92.23}{17.13}{***} & \mstd{25.70}{1.43}{***} & \mstd{0.804}{0.050}{***} & \mstd{72.43}{18.31}{***} & \mstd{28.14}{1.71}{***} & \mstd{0.825}{0.085}{***} & \mstd{99.17}{20.31}{***} & \mstd{25.43}{1.38}{***} & \mstd{0.781}{0.079}{***} \\
Brain/Pelvis & \bmstd{91.94}{27.30}{} & \bmstd{26.59}{2.34}{} & \bmstd{0.854}{0.038}{} & \mstd{117.23}{28.45}{***} & \mstd{24.44}{1.67}{***} & \mstd{0.822}{0.041}{***} & \mstd{96.67}{26.72}{***} & \mstd{26.26}{2.16}{***} & \mstd{0.842}{0.061}{***} & \mstd{121.39}{28.18}{***} & \mstd{24.31}{1.60}{***} & \mstd{0.811}{0.062}{***} \\
\bottomrule
\end{tabular}

\end{table*}
\renewcommand{\arraystretch}{1}

\begin{table*}[h!]
\centering
\scriptsize
\setlength{\tabcolsep}{2pt}
\renewcommand{\arraystretch}{1.25}
\caption{Quantitative results for Task 2 (supervised cross-validation) across anatomical regions. Mean $\pm$ standard deviation of commonly used image similarity metrics (MAE, PSNR, SSIM) are reported. Column groups indicate the registration method used during training (IMPACT or ELX), while subgroups correspond to the registration method used for evaluation. All regions correspond to in-distribution data used during training. Statistical comparisons against IMPACT/IMPACT configuration follow the patient-level protocol described in Section~\ref{sec:statistics}.
}
\label{tab:task2_cv_region_metrics_in_distribution}
\begin{tabular}{lcccccccccccc}
\toprule
 & \multicolumn{6}{c}{IMPACT} & \multicolumn{6}{c}{ELX} \\
\cmidrule(lr){2-7} \cmidrule(lr){8-13}
Region & \multicolumn{3}{c}{IMPACT} & \multicolumn{3}{c}{ELX} & \multicolumn{3}{c}{IMPACT} & \multicolumn{3}{c}{ELX} \\
\cmidrule(lr){2-4} \cmidrule(lr){5-7} \cmidrule(lr){8-10} \cmidrule(lr){11-13}
 & MAE & PSNR & SSIM & MAE & PSNR & SSIM & MAE & PSNR & SSIM & MAE & PSNR & SSIM \\
\midrule
AB & \bmstd{55.30}{14.14}{} & \bmstd{31.17}{2.40}{} & \bmstd{0.928}{0.023}{} & \mstd{67.10}{13.71}{***} & \mstd{28.96}{1.65}{***} & \mstd{0.909}{0.030}{***} & \mstd{62.96}{13.70}{***} & \mstd{29.88}{1.79}{***} & \mstd{0.910}{0.031}{***} & \mstd{59.08}{10.64}{} & \mstd{30.12}{1.57}{*} & \mstd{0.920}{0.025}{**} \\
HN & \bmstd{64.84}{14.73}{} & \bmstd{30.20}{2.07}{} & \bmstd{0.950}{0.019}{} & \mstd{77.18}{17.73}{***} & \mstd{28.33}{1.94}{***} & \mstd{0.935}{0.024}{***} & \mstd{71.74}{16.89}{***} & \mstd{29.40}{2.09}{***} & \mstd{0.938}{0.025}{***} & \mstd{68.55}{13.92}{**} & \mstd{29.29}{1.89}{**} & \mstd{0.943}{0.020}{***} \\
TH & \bmstd{55.40}{15.43}{} & \bmstd{32.21}{2.69}{} & \bmstd{0.930}{0.018}{} & \mstd{62.41}{12.66}{**} & \mstd{30.15}{1.95}{***} & \mstd{0.913}{0.028}{***} & \mstd{61.72}{16.16}{***} & \mstd{31.02}{2.45}{***} & \mstd{0.912}{0.023}{***} & \mstd{55.43}{12.51}{} & \mstd{31.32}{2.15}{*} & \mstd{0.925}{0.024}{} \\
AB/HN/TH & \bmstd{58.76}{15.47}{} & \bmstd{31.16}{2.53}{} & \bmstd{0.937}{0.022}{} & \mstd{69.17}{16.24}{***} & \mstd{29.13}{2.01}{***} & \mstd{0.920}{0.030}{***} & \mstd{65.70}{16.36}{***} & \mstd{30.08}{2.24}{***} & \mstd{0.921}{0.030}{***} & \mstd{61.28}{13.72}{**} & \mstd{30.22}{2.07}{***} & \mstd{0.930}{0.025}{***} \\
\bottomrule
\end{tabular}

\end{table*}
\renewcommand{\arraystretch}{1}

Across regions, the detailed results support the aggregate pattern reported in Table~\ref{tab:registration_consistency_aggregate}: performance is generally highest when the training and evaluation registration conventions are the same. The magnitude of the effect varies by anatomy, reflecting differences in deformation complexity and local image gradients.

\subsection{SAM-based supervision}
\label{app:sam_results}

Tables~\ref{tab:task1_cv_impact_mae_vs_sam_combined} and~\ref{tab:task2_cv_impact_mae_vs_sam_combined} report the complete Dice and SSIM results for models trained with MAE, VGG-based perceptual, or SAM-based perceptual supervision. Dice is obtained from an independent TotalSegmentator evaluation and therefore does not reuse the SAM representation employed during training.

\renewcommand{\arraystretch}{1}
\begin{table*}[h!]
\centering
\scriptsize
\setlength{\tabcolsep}{3pt}
\renewcommand{\arraystretch}{1.2}
\caption{Quantitative results for Task 1 using IMPACT-based training, comparing models trained with MAE, VGG-based perceptual, and SAM-based losses. Dice evaluates downstream segmentation performance, while SSIM measures image similarity between synthesized CT and reference CT. Values are reported as mean $\pm$ standard deviation across matched patients. Statistical comparisons against SAM follow the patient-level protocol described in Section~\ref{sec:statistics}.
}
\label{tab:task1_cv_impact_mae_vs_sam_combined}
\begin{tabular}{lcccccc}
\toprule
 & \multicolumn{2}{c}{MAE} & \multicolumn{2}{c}{VGG} & \multicolumn{2}{c}{SAM} \\
\cmidrule(lr){2-3} \cmidrule(lr){4-5} \cmidrule(lr){6-7}
Region & Dice & SSIM & Dice & SSIM & Dice & SSIM \\
\midrule
AB & \mstd{0.737}{0.044}{***} & \bmstd{0.909}{0.022}{***} & \mstd{0.737}{0.046}{***} & \mstd{0.904}{0.023}{} & \bmstd{0.777}{0.043}{} & \mstd{0.905}{0.022}{} \\
HN & \mstd{0.718}{0.067}{} & \bmstd{0.931}{0.027}{***} & \mstd{0.713}{0.067}{**} & \mstd{0.918}{0.033}{} & \bmstd{0.731}{0.060}{} & \mstd{0.914}{0.037}{} \\
TH & \mstd{0.679}{0.041}{***} & \bmstd{0.941}{0.018}{***} & \mstd{0.687}{0.038}{***} & \mstd{0.934}{0.020}{**} & \bmstd{0.706}{0.031}{} & \mstd{0.938}{0.020}{} \\
AB/HN/TH & \mstd{0.711}{0.057}{***} & \bmstd{0.927}{0.026}{***} & \mstd{0.712}{0.055}{***} & \mstd{0.919}{0.029}{} & \bmstd{0.738}{0.055}{} & \mstd{0.919}{0.030}{} \\
\addlinespace[0.25em]
\hline
\hline
\addlinespace[0.25em]
Brain & \mstd{0.778}{0.121}{**} & \mstd{0.858}{0.021}{} & \mstd{0.813}{0.100}{} & \bmstd{0.859}{0.019}{} & \bmstd{0.816}{0.089}{} & \mstd{0.857}{0.022}{} \\
Pelvis & \mstd{0.715}{0.120}{***} & \bmstd{0.850}{0.051}{} & \mstd{0.725}{0.115}{***} & \mstd{0.844}{0.046}{*} & \bmstd{0.795}{0.048}{} & \bmstd{0.850}{0.048}{} \\
Brain/Pelvis & \mstd{0.749}{0.125}{***} & \bmstd{0.854}{0.038}{} & \mstd{0.772}{0.116}{***} & \mstd{0.852}{0.035}{} & \bmstd{0.806}{0.074}{} & \bmstd{0.854}{0.036}{} \\
\addlinespace[0.25em]
\hline
\hline
\addlinespace[0.25em]
Ext-T2 & \mstd{0.665}{0.063}{***} & \mstd{0.899}{0.025}{*} & \mstd{0.672}{0.059}{***} & \bmstd{0.904}{0.023}{***} & \bmstd{0.725}{0.047}{} & \mstd{0.893}{0.032}{} \\
\bottomrule
\end{tabular}

\end{table*}

\begin{table*}[h!]
\centering
\scriptsize
\setlength{\tabcolsep}{3pt}
\renewcommand{\arraystretch}{1.2}
\caption{
Quantitative results for Task 2 using IMPACT-based training, comparing models trained with MAE, VGG-based perceptual, and SAM-based losses. Dice evaluates downstream segmentation performance, while SSIM measures image similarity between synthesized CT and reference CT. Values are reported as mean $\pm$ standard deviation. Statistical comparisons against SAM follow the patient-level protocol described in Section~\ref{sec:statistics}. $AB_{\mathrm{sim}}$, $HN_{\mathrm{sim}}$, and $TH_{\mathrm{sim}}$ correspond to the registration-free simulated CBCT evaluations.
}
\label{tab:task2_cv_impact_mae_vs_sam_combined}
\begin{tabular}{lcccccc}
\toprule
 & \multicolumn{2}{c}{MAE\_IMPACT} & \multicolumn{2}{c}{VGG\_IMPACT} & \multicolumn{2}{c}{SAM\_IMPACT} \\
\cmidrule(lr){2-3} \cmidrule(lr){4-5} \cmidrule(lr){6-7}
Region & Dice & SSIM & Dice & SSIM & Dice & SSIM \\
\midrule
AB & \mstd{0.642}{0.086}{***} & \bmstd{0.928}{0.023}{***} & \mstd{0.640}{0.083}{***} & \mstd{0.925}{0.024}{} & \bmstd{0.670}{0.086}{} & \mstd{0.925}{0.024}{} \\
HN & \bmstd{0.720}{0.078}{***} & \bmstd{0.950}{0.019}{***} & \mstd{0.697}{0.083}{***} & \mstd{0.943}{0.022}{} & \mstd{0.708}{0.076}{} & \mstd{0.943}{0.022}{} \\
TH & \mstd{0.718}{0.076}{} & \bmstd{0.930}{0.018}{***} & \mstd{0.709}{0.076}{***} & \mstd{0.928}{0.019}{} & \bmstd{0.720}{0.077}{} & \mstd{0.927}{0.020}{} \\
AB/HN/TH & \mstd{0.695}{0.088}{} & \bmstd{0.937}{0.022}{***} & \mstd{0.683}{0.086}{***} & \mstd{0.932}{0.023}{} & \bmstd{0.700}{0.082}{} & \mstd{0.932}{0.023}{} \\
\addlinespace[0.25em]
\hline
\hline
\addlinespace[0.25em]
$AB_{\mathrm{sim}}$ & \mstd{0.709}{0.085}{***} & \mstd{0.927}{0.016}{***} & \mstd{0.723}{0.077}{***} & \mstd{0.921}{0.016}{***} & \bmstd{0.764}{0.079}{} & \bmstd{0.932}{0.015}{} \\
$HN_{\mathrm{sim}}$ & \mstd{0.734}{0.065}{***} & \mstd{0.901}{0.029}{***} & \mstd{0.750}{0.061}{*} & \mstd{0.902}{0.028}{***} & \bmstd{0.755}{0.063}{} & \bmstd{0.907}{0.028}{} \\
$TH_{\mathrm{sim}}$ & \mstd{0.722}{0.120}{***} & \mstd{0.885}{0.062}{***} & \mstd{0.741}{0.114}{***} & \mstd{0.888}{0.057}{***} & \bmstd{0.758}{0.110}{} & \bmstd{0.892}{0.059}{} \\
\bottomrule
\end{tabular}

\end{table*}
\renewcommand{\arraystretch}{1}

The region-wise results show that SAM-based supervision generally improves downstream Dice, particularly in OOD regions, although SSIM
does not always improve. This difference supports the conclusion that anatomical preservation and voxel-wise agreement with a registered reference provide complementary information.

\subsection{Perceptual and ensemble results}
\label{app:perceptual_results}

Tables~\ref{tab:task1_elx_fold_vs_cv_mae_median_perceptual} and~\ref{tab:task2_elx_fold_vs_cv_mae_median_perceptual} provide the complete regional comparison between MAE- and SAM-trained models. Mean CV denotes patient-level performance averaged over the five individual fold models, whereas CV denotes evaluation after voxel-wise averaging of the five predictions.

\begin{table*}[h!]
\centering
\scriptsize
\setlength{\tabcolsep}{3pt}
\renewcommand{\arraystretch}{1.2}
\caption{
Quantitative results for Task 1 using models trained with ELX and evaluated with ELX, comparing MAE- and SAM-based training losses. “CV” denotes the ensemble prediction obtained by averaging the outputs of the five cross-validation models (CV$_0$–CV$_4$), while “Mean CV” corresponds to the average performance of the individual models. Performance is reported using MAE, $d_{\mathrm{SAM}}$, and LPIPS (mean $\pm$ standard deviation) computed on matched patients. $d_{\mathrm{SAM}}$ and LPIPS values are scaled by a factor of 100 for readability.
}
\label{tab:task1_elx_fold_vs_cv_mae_median_perceptual}
\begin{tabular}{lcccccccccccc}
\toprule
Region & \multicolumn{6}{c}{Mean CV} & \multicolumn{6}{c}{CV} \\
\cmidrule(lr){2-7} \cmidrule(lr){8-13}
 & \multicolumn{3}{c}{MAE} & \multicolumn{3}{c}{SAM} & \multicolumn{3}{c}{MAE} & \multicolumn{3}{c}{SAM} \\
\cmidrule(lr){2-4} \cmidrule(lr){5-7} \cmidrule(lr){8-10} \cmidrule(lr){11-13}
 & MAE & SAM & LPIPS & MAE & SAM & LPIPS & MAE & SAM & LPIPS & MAE & SAM & LPIPS \\
\midrule
AB & \mstd{71.30}{14.44}{} & \mstd{32.54}{5.94}{} & \mstd{12.67}{3.42}{} & \mstd{74.00}{13.84}{} & \bmstd{25.64}{4.38}{} & \mstd{10.66}{2.79}{} & \bmstd{67.73}{14.25}{} & \mstd{32.46}{6.07}{} & \mstd{12.49}{3.35}{} & \mstd{69.35}{14.04}{} & \mstd{26.70}{4.50}{} & \bmstd{10.66}{2.81}{} \\
HN & \mstd{83.51}{24.76}{} & \mstd{14.54}{2.37}{} & \mstd{3.46}{0.63}{} & \mstd{99.88}{33.98}{} & \bmstd{11.63}{1.76}{} & \mstd{3.11}{0.54}{} & \bmstd{78.98}{23.41}{} & \mstd{14.76}{2.37}{} & \mstd{3.41}{0.64}{} & \mstd{93.66}{33.16}{} & \mstd{11.96}{1.82}{} & \bmstd{3.05}{0.55}{} \\
TH & \mstd{58.57}{10.54}{} & \mstd{25.25}{5.66}{} & \mstd{8.66}{2.88}{} & \mstd{62.60}{10.62}{} & \bmstd{19.35}{4.77}{} & \mstd{7.23}{2.33}{} & \bmstd{55.30}{10.14}{} & \mstd{25.24}{5.71}{} & \mstd{8.52}{2.86}{} & \mstd{58.22}{10.12}{} & \mstd{20.50}{5.19}{} & \bmstd{7.18}{2.41}{} \\
AB/HN/TH & \mstd{70.75}{20.17}{} & \mstd{24.27}{8.83}{} & \mstd{8.34}{4.56}{} & \mstd{78.26}{26.66}{} & \bmstd{18.99}{6.88}{} & \mstd{7.06}{3.72}{} & \bmstd{66.98}{19.27}{} & \mstd{24.31}{8.77}{} & \mstd{8.22}{4.49}{} & \mstd{73.21}{25.84}{} & \mstd{19.85}{7.26}{} & \bmstd{7.03}{3.77}{} \\
\addlinespace[0.25em]
\hline
\hline
Ext-T2 & \mstd{104.39}{16.92}{} & \mstd{37.43}{4.39}{} & \mstd{13.82}{2.95}{} & \mstd{96.36}{15.38}{} & \bmstd{33.71}{3.86}{} & \mstd{12.00}{2.51}{} & \mstd{97.43}{14.83}{} & \mstd{37.30}{4.17}{} & \mstd{13.25}{2.95}{} & \bmstd{90.14}{14.09}{} & \mstd{35.93}{3.93}{} & \bmstd{11.79}{2.58}{} \\
\bottomrule
\end{tabular}

\end{table*}
\renewcommand{\arraystretch}{1}

\begin{table*}[h!]
\centering
\scriptsize
\setlength{\tabcolsep}{3pt}
\renewcommand{\arraystretch}{1.2}
\caption{
Quantitative results for Task 2 using models trained with ELX and evaluated with ELX, comparing MAE- and SAM-based training losses. “CV” denotes the ensemble prediction obtained by averaging the outputs of the five cross-validation models (CV$_0$–CV$_4$), while “Mean CV” corresponds to the average performance of the individual models. Performance is reported using MAE, $d_{\mathrm{SAM}}$, and LPIPS (mean $\pm$ standard deviation). $d_{\mathrm{SAM}}$ and LPIPS values are scaled by a factor of 100 for readability.
}
\label{tab:task2_elx_fold_vs_cv_mae_median_perceptual}
\begin{tabular}{lcccccccccccc}
\toprule
Region & \multicolumn{6}{c}{Mean CV} & \multicolumn{6}{c}{CV} \\
\cmidrule(lr){2-7} \cmidrule(lr){8-13}
 & \multicolumn{3}{c}{MAE} & \multicolumn{3}{c}{SAM} & \multicolumn{3}{c}{MAE} & \multicolumn{3}{c}{SAM} \\
\cmidrule(lr){2-4} \cmidrule(lr){5-7} \cmidrule(lr){8-10} \cmidrule(lr){11-13}
 & MAE & SAM & LPIPS & MAE & SAM & LPIPS & MAE & SAM & LPIPS & MAE & SAM & LPIPS \\
\midrule
AB & \mstd{62.55}{10.57}{} & \mstd{21.83}{5.95}{} & \mstd{7.82}{2.95}{} & \mstd{66.39}{11.88}{} & \bmstd{17.57}{4.06}{} & \mstd{6.69}{2.35}{} & \bmstd{59.08}{10.64}{} & \mstd{22.19}{6.32}{} & \mstd{7.73}{2.99}{} & \mstd{62.30}{12.00}{} & \mstd{18.10}{4.75}{} & \bmstd{6.60}{2.45}{} \\
HN & \mstd{72.35}{13.72}{} & \mstd{11.48}{2.20}{} & \mstd{2.48}{0.70}{} & \mstd{78.71}{14.05}{} & \bmstd{9.40}{1.53}{} & \mstd{2.22}{0.61}{} & \bmstd{68.55}{13.92}{} & \mstd{11.75}{2.39}{} & \mstd{2.42}{0.71}{} & \mstd{74.24}{14.18}{} & \mstd{9.50}{1.64}{} & \bmstd{2.09}{0.61}{} \\
TH & \mstd{59.16}{12.89}{} & \mstd{20.09}{4.91}{} & \mstd{6.53}{2.31}{} & \mstd{62.46}{12.90}{} & \bmstd{15.63}{3.81}{} & \mstd{5.51}{1.85}{} & \bmstd{55.43}{12.51}{} & \mstd{20.51}{4.86}{} & \mstd{6.44}{2.28}{} & \mstd{58.05}{12.66}{} & \mstd{16.12}{3.90}{} & \bmstd{5.42}{1.88}{} \\
AB/HN/TH & \mstd{64.95}{13.77}{} & \mstd{17.54}{6.46}{} & \mstd{5.48}{3.15}{} & \mstd{69.52}{14.82}{} & \bmstd{13.99}{4.82}{} & \mstd{4.69}{2.58}{} & \bmstd{61.28}{13.72}{} & \mstd{17.88}{6.62}{} & \mstd{5.40}{3.15}{} & \mstd{65.18}{14.79}{} & \mstd{14.36}{5.18}{} & \bmstd{4.59}{2.63}{} \\
\addlinespace[0.25em]
\hline
\hline
\addlinespace[0.25em]
Sim-CBCT & \mstd{114.78}{46.81}{} & \mstd{20.87}{6.07}{} & \mstd{6.34}{2.87}{} & \mstd{109.23}{43.70}{} & \bmstd{17.94}{4.63}{} & \mstd{5.31}{2.32}{} & \mstd{111.88}{46.90}{} & \mstd{20.67}{6.10}{} & \mstd{6.15}{2.80}{} & \bmstd{106.33}{44.17}{} & \mstd{17.99}{5.00}{} & \bmstd{5.15}{2.34}{} \\
\bottomrule
\end{tabular}

\end{table*}
\renewcommand{\arraystretch}{1}

Across both tasks, SAM supervision improves $d_{\mathrm{SAM}}$ and LPIPS in the in-distribution regions while often increasing MAE. In the Ext-T2 and Sim-CBCT settings, it improves both perceptual and voxel-wise metrics. Ensemble averaging primarily reduces MAE, whereas its effect on perceptual distances is smaller, consistent with averaging reducing random intensity errors while potentially smoothing fine structures.

\end{document}